\documentclass[lettersize,journal]{IEEEtran}
\usepackage{array}
\usepackage[caption=false,font=normalsize,labelfont=sf,textfont=sf]{subfig}
\usepackage{textcomp}
\usepackage{stfloats}

\usepackage{cite}
\usepackage{amsmath,amssymb,amsfonts}
\usepackage{algorithmic}
\usepackage{graphicx}
\usepackage{textcomp}
\usepackage{xcolor}

\usepackage{booktabs}
\usepackage{enumitem}
\usepackage{float}
\usepackage{mathrsfs}
\usepackage{bbding}
\usepackage{url}
\usepackage{xcolor}
\usepackage{calligra}
\usepackage[linesnumbered,ruled,lined,boxed]{algorithm2e}
\usepackage{multirow}
\usepackage[misc]{ifsym}

\usepackage{manfnt}
\usepackage{fontawesome}

\newcommand{\bluecallig}[1]{{\scalebox{1.05}{\color{blue}{\textbf{\textit{#1}}}}}}
\newcommand{\redcallig}[1]{{\scalebox{1.05}{\color{red}{\textbf{\textit{#1}}}}}}
\newcommand{\blackcallig}[1]{{\scalebox{1.05}{\color{black}{\textbf{\textit{#1}}}}}}

\newcommand{\syx}[1]{{\color{black} #1}}
\newcommand{\revision}[1]{{\color{black} #1}}

\def\BibTeX{{\rm B\kern-.05em{\sc i\kern-.025em b}\kern-.08em
    T\kern-.1667em\lower.7ex\hbox{E}\kern-.125emX}}
\usepackage{balance}
\begin{document}
\title{\LARGE \bf
Learning 6-DoF Fine-grained Grasp Detection Based on \\ Part Affordance Grounding
}

\author{Yaoxian Song$^{\dag}$, Penglei Sun$^{\dag}$, Piaopiao Jin, Yi Ren,  Yu Zheng, Zhixu Li, Xiaowen Chu,\\ Yue Zhang, Tiefeng Li\textsuperscript{\Envelope}, and Jason Gu\textsuperscript{\Envelope}
\thanks{\dag~Equal contribution. \Envelope~Corresponding author.}
\thanks{Yaoxian Song, Piaopiao Jin and Tiefeng Li are with the Center for X-Mechanics, School of Aeronautics and Astronautics, Zhejiang University, Hangzhou, Zhejiang, China, Emails: 
    {\tt\small \{songyaoxian,piaopiaojin,litiefeng\}@zju.edu.cn.}}
\thanks{Penglei Sun and Xiaowen Chu are with the Information Hub, Data Science and Analytics Thrust, The Hong Kong University of Science and Technology (Guangzhou), Guangdong, China. Emails:
    {\tt\small psun012@connect.hkust-gz.edu.cn, xwchu@hkust-gz.edu.cn}.}
\thanks{Yi Ren is with the Advanced Manufacturing Lab, Huawei Technologies, Shenzhen, Guangdong, China. Email:
    {\tt\small even.renyi@huawei.com.}}
\thanks{Yu Zheng is with the Research Institute, UBTECH Robotics Inc., Shenzhen, Guangdong, China. Email:
    {\tt\small yu.zheng3@ubtrobot.com.}}
\thanks{Zhixu Li is with School of Information and School of Smart Governance, Renmin University of China, Beijing, China. Email: {\tt\small zhixuli@ruc.edu.cn.}}
\thanks{Yue Zhang is with the School of Engineering, Westlake University and the Institute of Advanced Technology, Westlake Institute for Advanced Study, Hangzhou, Zhejiang, China, Email: 
    {\tt\small zhangyue@westlake.edu.cn.}}
\thanks{Jason Gu is with the Department of Electrical and Computer Engineering, Dalhousie University, Halifax, NS Canada B3H 4R2. 
    Email: {\tt\small jason.gu@dal.ca.}}
}


\markboth{Journal of \LaTeX\ Class Files,~Vol.~18, No.~9, April~2025}%
{How to Use the IEEEtran \LaTeX \ Templates}

\maketitle

\begin{abstract}
Robotic grasping is a fundamental ability for a robot to interact with the environment. Current methods focus on how to obtain a stable and reliable grasping pose in object level, while little work has been studied on part (shape)-wise grasping which is related to fine-grained grasping and robotic affordance.
Parts can be seen as atomic elements to compose an object, which contains rich semantic knowledge and a strong correlation with affordance.
However, lacking a large part-wise 3D robotic dataset limits the development of part representation learning and downstream applications.
In this paper, we propose a new large \textbf{Lang}uage-guided \textbf{SH}ape gr\textbf{A}s\textbf{P}ing datas\textbf{E}t (named \underline{\textbf{LangSHAPE}}) to promote 3D part-level affordance and grasping ability learning. From the perspective of robotic cognition, we design a two-stage fine-grained robotic grasping framework (named \underline{\textbf{LangPartGPD}}), including a novel 3D part language grounding model and a part-aware grasp pose detection model, in which explicit language input from human or large language models (LLMs) could guide a robot to generate part-level 6-DoF grasping pose with textual explanation. Our method combines the advantages of human-robot collaboration and LLMs' planning ability using explicit language as a symbolic intermediate.
To evaluate the effectiveness of our proposed method, we perform 3D part grounding and fine-grained grasp detection experiments on both simulation and physical robot settings, following \syx{language instructions across different degrees of textual complexity}. 
Results show our method achieves competitive performance in 3D geometry fine-grained grounding, object affordance inference, and 3D part-aware grasping tasks. Our dataset and code are available on our project website~\url{https://sites.google.com/view/lang-shape}
\end{abstract}

\def\abstractname{Note to Practitioners}
\begin{abstract}
This paper was motivated by the problem of pragmatic 6 DoF robotic grasping for a robot in a real-world scenario. Different from existing work focusing on task-oriented solutions, our research attempts to consider fine-grained grasping as a general and essential modeling problem (i.e. \underline{Point 1}: which part to grasp; \underline{Point 2}: how to grasp the specific part). For that, our method fully considers the scalability and compatibility with subsequently developed functional modules. A two-stage loosely-coupled fine-grained grasping framework (LangPartGPD) is designed, including a 3D part language grounding model and a part-aware 6 DoF grasping pose detection model (\underline{Point 1}). The former is used to link the symbolic-level grasped part representation to the specific region of an object. The latter is to realize kinematic-level grasping point generation based on the constraint of parts (\underline{Point 2}). To realize the designed model, we propose a newly large-scale language-pointcloud-grasp dataset (LangSHAPE) to train fine-grained 3D part-level grounding and part-constrained grasping detection models. It is noted that our method is purely concerned with the \underline{target object}, \underline{grasped part}, and \underline{functionality to realize} to model the essential fine-grained graspability, which is an insufficiently explored direction currently and could promote task-level grasping performance. The symbolic-level description of grasping in the form of natural language provides an explainable decision and a user-friendly interface for human-robot collaboration. Experiments in simulations and real-robot settings show our method can effectively enable a robot to make a cognitive decision for fine-grained manipulation. Summarily, our work is a preliminary exploration to link symbolic-level cognitive space to kinematic-level control space directly. In future research, we will further combine more complicated cognitive knowledge from public multimodal knowledge graphs and multimodal LLMs, to build a general robotic cognitive decision system using LangPartGPD as the basic components.
\end{abstract}


\begin{figure}[t]
	\centering 
    \setlength{\belowcaptionskip}{-0.5cm}
	\includegraphics[width=\linewidth]{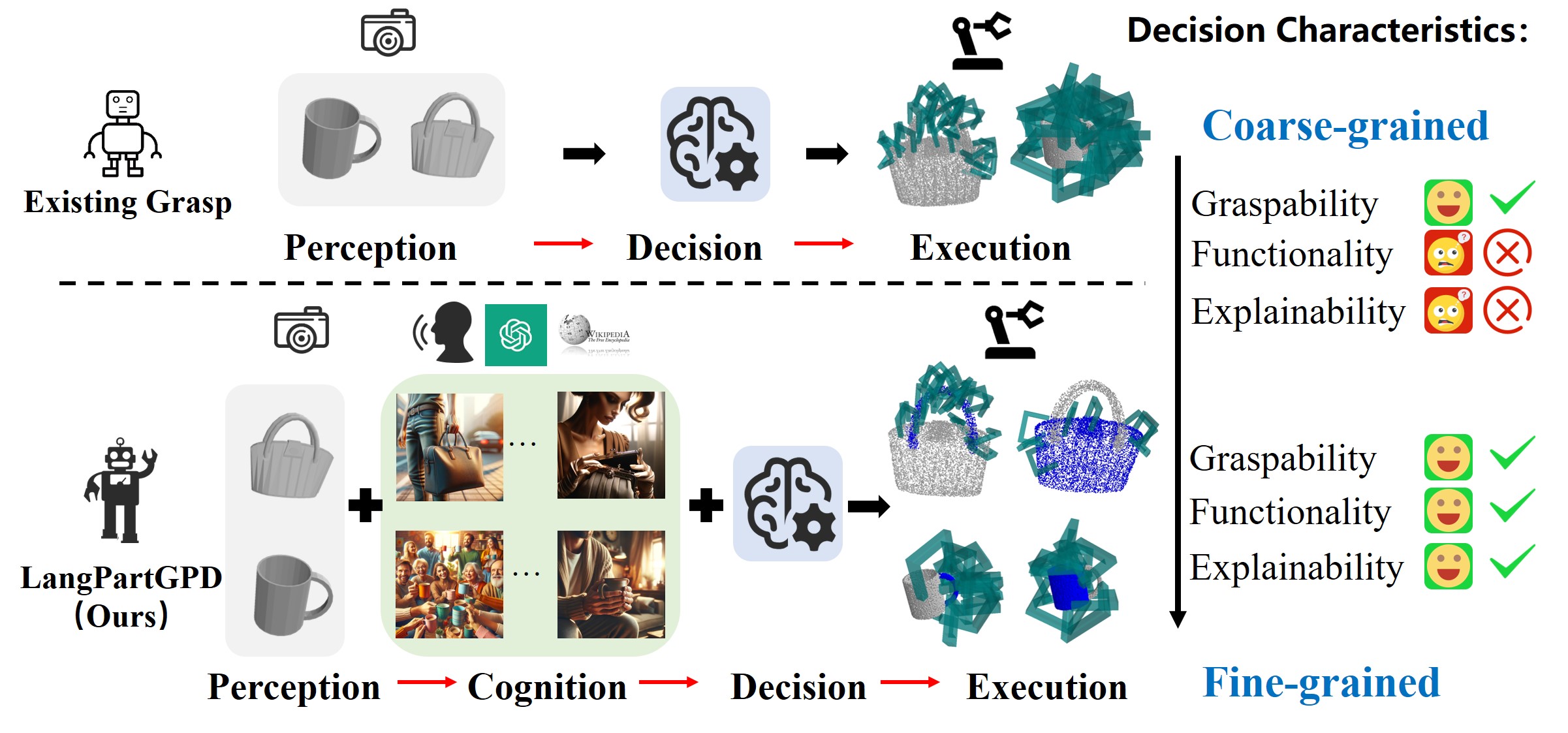}
	\caption{Illustration of 6-DoF grasp detection. Most of the existing work focuses on the graspability of grasp pose detection via visual perception, while \blackcallig{ignoring} grasp semantics for grasping functionality and explainability of decision. Our method attempts to overcome these limitations by introducing an intermediate process (i.e., 3D part language grounding) using explicit natural language from humans or LLMs.}
	\label{fig:palcement}
\end{figure}

\section{Introduction}
Fine-grained robotic manipulation can allow a robot to tackle human tasks by mimicking human hands in embodied AI~\cite{zhu2020dark}. Different from low-level manipulation in control (e.g., pick and place), fine-grained robotic manipulation not only has the abilities (e.g., graspability) but also considers additional details (e.g., which part to grasp; why to grasp this part), which reduces to affordance~\cite{gibson1977theory}. For example, \textit{when a human demonstrates the intention of drinking a cup of coffee and hopes a robot to bring his mug, the robot is expected to grasp the handle and not pollute the inside of the mug subliminally}. Behind that, the affordance of manipulated objects is needed, and how to empower low-level control policy with high-level concepts and knowledge represented by human symbolic language has received increasing attention in the robotics community~\cite{ahn2022can}.

\begin{table*}[t]
\centering
\caption{Comparison of publicly available grasp datasets. The label of grasping and text reference is generated either manually(\HandPencilLeft), by physical simulation(\faPlayCircle), or analytical method($f$).}
\resizebox{0.8\linewidth}{!}{
\begin{tabular}{@{}|c|c|c|c|c|c|c|c|c|c|c|@{}}
\toprule
\textbf{Dataset}                                               & \textbf{\begin{tabular}[c]{@{}c@{}}Plannar\\ /3D\end{tabular}} & \textbf{\begin{tabular}[c]{@{}c@{}}Part\\ -aware\end{tabular}} & \textbf{\begin{tabular}[c]{@{}c@{}}Obser-\\ vations\end{tabular}} & \textbf{Labels} & \textbf{Grasps} & \textbf{\begin{tabular}[c]{@{}c@{}}Objects\\ (Cat.)\end{tabular}} & \textbf{\begin{tabular}[c]{@{}c@{}}Part\\ (Cat.)\end{tabular}} & \textbf{\begin{tabular}[c]{@{}c@{}}Grasps\\ per part\end{tabular}} & \textbf{Reference} & \textbf{\begin{tabular}[c]{@{}c@{}}Reference\\ per part\end{tabular}} \\ \midrule
Cornell~\cite{jiang2011efficient}                                                        & $\square$                                                             & \XSolidBrush                                                              & Real                                                                 & \HandPencilLeft               & 8k               & 240                & -                                                              & -                                                                  & -                  & -                                                                     \\ \midrule
Jacquard~\cite{depierre2018jacquard}                                                       & $\square$                                                            & \XSolidBrush                                                              & Sim                                                                 & \faPlayCircle               & 1.1M               & 11k                & -                                                              & -                                                                  & -                  & -                                                                     \\ \midrule
VMRD~\cite{zhang2019roi}                                                           & $\square$                                                              & \XSolidBrush                                                              & Real                                                                 & \HandPencilLeft               & 100k               & 15k (31)                & -                                                              & -                                                                  & -                  & -                                                                     \\ \midrule
Dex-Net~\cite{mahler2017dex}                                                        & \mancube                                                            & \XSolidBrush                                                              & Sim                                                                 & $f$               & 6.7M               & 1500 (50)               & -                                                              & -                                                                  & -                  & -                                                                     \\ \midrule
GraspNet~\cite{fang2020graspnet}                                                       & \mancube                                                              & \XSolidBrush                                                              & S+R                                                                 & $f$               & 1.1B               & 88                & -                                                              & -                                                                 & -                  & -                                                                     \\ \midrule
\begin{tabular}[c]{@{}c@{}}Affordance\\ -language~\cite{nguyen20}\end{tabular} & \mancube                                                              & \XSolidBrush                                                              & Real                                                                 & \HandPencilLeft                & -               & 216                & -                                                              & -                                                                  & 655                  & -                                                                     \\ \midrule
ACRONYM~\cite{eppner2021acronym}                                                        & \mancube                                                              & \XSolidBrush                                                              & Sim                                                                 & \faPlayCircle               & 17.7M               & 8872 (262)                & -                                                              & -                                                                  & -                  & -                                                                     \\ \midrule
\textbf{LangSHAPE(ours)}                                                    & \mancube                                                              & \checkmark                                                              & Sim                                                                 & \faPlayCircle, $f$,\HandPencilLeft               & 2.5M               & 16.6k (16)                & 42.1k (35)                                                              & 60                                                                  & 1.38M                  & 99                                                                     \\ \bottomrule
\end{tabular}
}
\label{tab:dataset_compare}
\end{table*}

Thanks to recent progress in artificial intelligence fields (e.g., computer vision and natural language processing), vision-based robotic manipulation methods have achieved great success.
Many efforts are made by researchers to explore steady, dexterous, reliable grasping policy~\cite{newbury2023deep}. 
Most of the grasping detection methods can be divided into object-agnostic method and object-aware method~\cite{duan2021robotics}. For object-agnostic grasping, researchers consider pixel-wise image regression~\cite{lenz2015deep,morrison2018closing,wang2022transformer,song2022deep,yu2022skgnet} and sampling-based methods~\cite{chu2018real,zhang2019roi,ten2017grasp,mahler2017dex,liang2019pointnetgpd,fang2020graspnet,sundermeyer2021contact,peng2021self,laili2022custom,fang2023anygrasp} in an image or point cloud to realize 2D planar and spatial 6-DoF grasping detection. For object-aware grasping, object detection and segmentation are widely used to realize single or cluttered grasping detection~\cite{vohra2019real,zhang2019multi,yu2020robotic,zhang2021invigorate,zeng2021ppr,song2022human,kim2022dsqnet}, in which object category, shape, and spatial relation are usually considered to improve grasping performance in stability and collision free task. In particular, affordance is a key attribute for an autonomous robotic agent to effectively interact with novel objects, such as robotic manipulation operation~\cite{ardon2020affordances}. It can reduce the entire object in the sampling space to the area where the grasp is most likely to succeed~\cite {ardon2019learning}. Image-based affordance \cite{chu2019toward,xu2021affordance,qian2020grasp,mandikal2021learning} and point cloud-based affordance~\cite{tekdenaffordance,murali2021same,chen2022learning,nguyen2023language} are extracted to make grasping detection based on knowledge of human grasping habits. However, firstly, for robotic grasping detection, as shown in Fig.~\ref{fig:palcement}, most of the existing methods (e.g., object-agnostic or object-aware methods) focus on object-level grasping to achieve grasp stability. A more fine-grained grasp considering the affordance of an object based on parts composition has not been explored sufficiently. Secondly, for affordance in robotics, although there exists some research on task-oriented manipulation~\cite{tang2023graspgpt} using affordance as a new physical and geometric property of objects, they utilize object-level affordance to complete specific tasks with limited scale of objects missing the generalization and relation of part-level semantic with graspability. Furthermore, most affordance detection work depends on 2D image technology as pixel segmentation which projects to 3D space, while direct 3D part-level affordance detection technology has not been developed, especially using explicit visual-language pairs as input with open-world human knowledge, which could provide convincing and valid support for robotic action to guarantee the explainability of decisions.

To overcome the challenges in existing work on part-level fine-grained grasping detection with direct 3D affordance reasoning, we attempt to explore solutions from \textbf{(1) novel large dataset for affordance} and \textbf{(2) grasping detection and fine-grained 6-DoF grasping detection modeling}.
Specifically, we first propose a large language-guided shape grasp dataset (named \textbf{LangSHAPE}) for 6-DoF fine-grained robotic grasping shown in Fig.~\ref{fig:palcement}. We utilize symbolic knowledge about affordance from commonsense knowledge graph and natural language corpus, object-part-level point cloud, and 6-DoF grasping detection theory to make a large-scale fine-grained language-pointcloud-grasp dataset, shown in Fig.~\ref{fig:dataset_gen}. More than $1.38$ million sentences are generated via ChatGPT~\cite{ChatGPTO40:online} based on $5$ kinds ($99$ in total) of templates. Comparisons with typical grasping datasets are given in TABLE~\ref{tab:dataset_compare}. 
Secondly, we consider fine-grained 6-DoF grasping based on the part of an object by decoupling grasping semantic inference and grasping detection. We first propose a 3D part language grounding model to link the symbolic part concepts with 3D part geometry in the form of point cloud. The part concepts refer to the composition relationship of object, geometrical shape, affordance of object and part, and intention within human-robot interaction. In the following, a unified sampling-based part-aware 6-DoF grasping detection framework is introduced based on 3D part language grounding, in which we use 3D part language grounding to constrain sampling region for obtaining semantic-related grasping candidates~\cite{ten2017grasp}. The whole pipeline is shown in Fig.~\ref{fig:network_struture}.

Our main contributions can be summarized as follows:

\begin{enumerate}
    \item We consider part-wise affordance for spatial 6-DoF robotic grasping by human-robot knowledge sharing using explicit symbolic grounding. Human-level cognitive knowledge in natural language and structured knowledge graph are adopted to improve intrinsic understanding of objects and fine-grained graspability for a robot arm in 3D space with a language decision basis.

    \item A large-scale language-pointcloud-grasp dataset is constructed covering affordance of grasped targets and human intention, named \textbf{LangSHAPE}. The dataset is the first of its kind to provide fine-grained spatial grasping labels with millions of natural language instructions. 

    \item We propose a part-level 6-DoF grasp point detection model, named \textbf{LangPartGPD}. It is a pragmatic framework including a 3D part language grounding model and a part-aware grasp detection model. The former links the 3D point cloud with abstract affordance or intention concepts. The latter utilizes an affordance-guided sampling policy to realize efficient fine-grained grasp operation.

    \item \revision{We evaluate the proposed method using language instructions with multi-level textual complexities. Corrupted language instructions are used to test the inference capability of the grounding model, and our method achieves over $\mathbf{10\%}$ improvement on average compared to existing work in Part avg IoU (known\_all). In simulation experiments using MuJoCo~\cite{todorov2012mujoco}, our method outperforms existing methods by more than $\mathbf{36\%}$ success rate in part-specific grasping. In physical experiments using Kionva and Intel RealSense cameras, baseline grasping detection models intervened by our proposed 3D part grounding can achieve at least $\mathbf{11.8\%}$ improvements in part-agnostic success rate, and more than $\mathbf{67.5\%}$ part-specific grasping
    success rate in part\_unknown instruction setting. All results indicate the effectiveness and superiority of our grasping detection models using affordance-based part grounding.}
\end{enumerate}

\begin{figure}[t]
	\centering 
 	\setlength{\belowcaptionskip}{-0.5cm}
	\includegraphics[width=\linewidth]{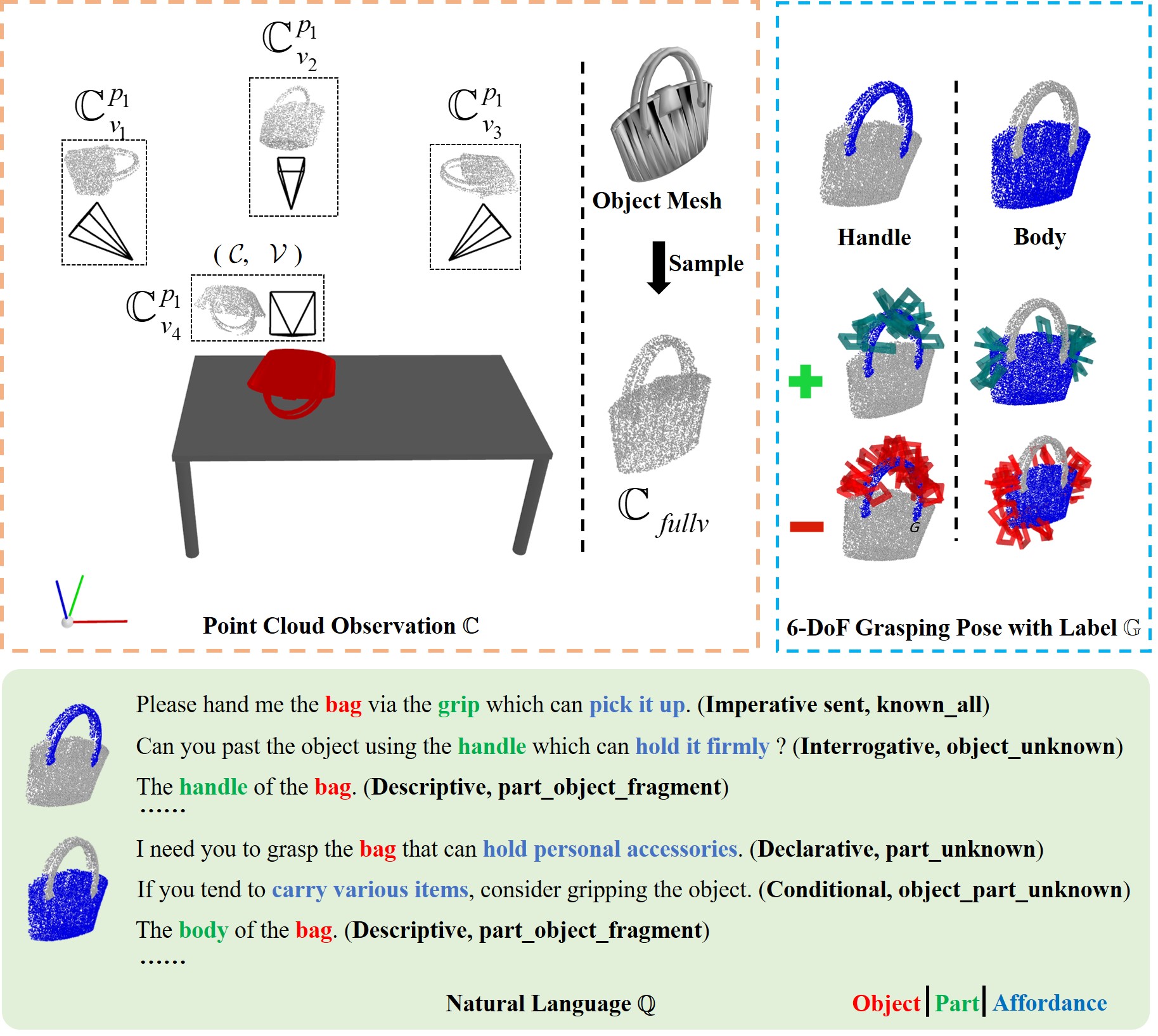}
	\caption{An example of LangSHAPE with point cloud observation $\mathbb{C}$ under 1st time random placement, 6-DoF grasping pose with label $\mathbb{G}$, and natural language $\mathbb{Q}$ about part, object and affordance for grasping.}
	\label{fig:example_langshape}
\end{figure}

\section{Related work}
\subsection{Language Grounding in Robotic Manipulation}
Visual language grounding is to localize a region described by a given referring expression~\cite{chen2020scanrefer}. For 2D computer vision, the localized region is usually a bounding box~\cite{mees2020composing,chen2021joint,zhang2021invigorate,song2022human,qianthinkgrasp2024,huang2024manipvqa,tong2024oval,tziafastowards} or heatmap~\cite{qian2024affordancellm,li2024one}, in which the robots are usually set up on the tabletop to realize planar grasping. However, this method is sensitive to occlusion because of the use of the coarse-grained bounding box rather than the pixel-wise segmentation mask.
For 3D computer vision, most of the grounding work~\cite{chen2020scanrefer,zhao20213dvg,zhang2023multi3drefer,wang2024find,liu2024dragon,xuvlm2024} are scene-level to link language concepts with 3D observation for navigation tasks. A closely related work by~\cite{thomason2022language,song2024scene} studied to reason visual and non-visual language about 3D objects, which is a binary classification task rather than real grounding to 3D geometry. On the other hand, to analyze the semantics of the language, although there are several language grounding methods used in robotic grasping~\cite{mees2020composing,chen2021joint,zhang2021invigorate,ahn2022can,song2022human,qianthinkgrasp2024}, most of them consider direct command (e.g., abstract action) or scene understanding with a spatial relationship in object wise~\cite{huang2024grounded} and object with affordance in part wise has not been investigated. Different from the existing methods, we attempt to consider part-wise grounding language on point cloud in a 6-DoF grasping task to realize fine-grained multiple abstract concepts linking with real-world objects.

\subsection{Affordance in Robotic Grasping Detection}
The possible action an agent could make to interact with the object in the environment and the functionality is a permanent property of an object independent of the characteristics of the user~\cite{gibson1979theory}, which is the core idea of affordance theory. Affordances in robotic tasks are more likely to be associated with features such as the geometrical shape and texture of an object than they are with its object class~\cite{bohg2010learning,ardon2020affordances}, which have envisioned affordances perceived by robots as an imitation of how humans might learn about affordances during their development as infants. It provides a more approximate intermediate representation for object prediction than object classification during scene understanding. Affordance segmentation and reasoning are two primary tasks for robot scene understanding and decision.
To detect grasp point in pixel level, Vahrenkamp et al.~\cite{vahrenkamp2016part}, Chu et al.~\cite{chu2019learning} and Wang et al.~\cite{wang2023task} propose an affordance segmentation via synthetic images to realize planar grasping based on part affordance in the conventional mask. With the development of multimodal large language models, Tong et al.~\cite{tong2024oval} design a prompt-based approach for open-vocabulary affordance localization in RGB-D images using GPT-4~\cite{achiam2023gpt}.
Furthermore, Xu et al.~\cite{xu2021affordance} introduce an affordance keypoint detection by providing structured predictions regarding the position, direction, and extent. Li et al.~\cite{li2025language} adopt visual foundation models (VFMs)~\cite{sun2023going} and the Segment Anything Mode (SAM)~\cite{kirillov2023segment} to realize visual affordance grounding in pixel level.
Recently, with the advent of 3D object part point cloud dataset~\cite{kim2014semantic,deng20213d,xu2022partafford}, point-wise spatial grasping detection with affordance based on point cloud is proposed~\cite{ardon2019learning,chen2022learning}. \cite{ahn2022can,tang2023graspgpt} adopt affordance reasoning by large language models to make decisions of object selection or tool usage for complex indoor manipulation tasks. Compared with existing research, pixel-based affordance cannot be used by 6-DoF grasping methods directly, which greatly diminishes its practical value. The training set exposes deficiencies in the scale, granularity, and richness of corpus content. Whether \cite{myers2015affordance,nguyen2017object,sawatzky2017weakly,do2018affordancenet} 2D affordance dataset or 3D dataset~\cite{deng20213d}, they have limited scale of visual data and use handcrafted labels to description affordance. It is hard to guarantee generalization and robustness in real-world applications. Although some research uses open-world method~\cite{nguyen2023open}, such as SAM~\cite{kirillov2023segment} to alleviate deficiencies, we attempt to solve problems starting from first principles by constructing novel 3D part-level point cloud dataset with ego-centric language description about affordance of the object and part. For affordance reasoning of robotic manipulation, we propose a new task named 3D language part grounding to directly link perception data with cognitive concepts in physical interactive or contact surfaces.

\begin{figure*}[]
	\centering 
	\includegraphics[width=0.95\linewidth]{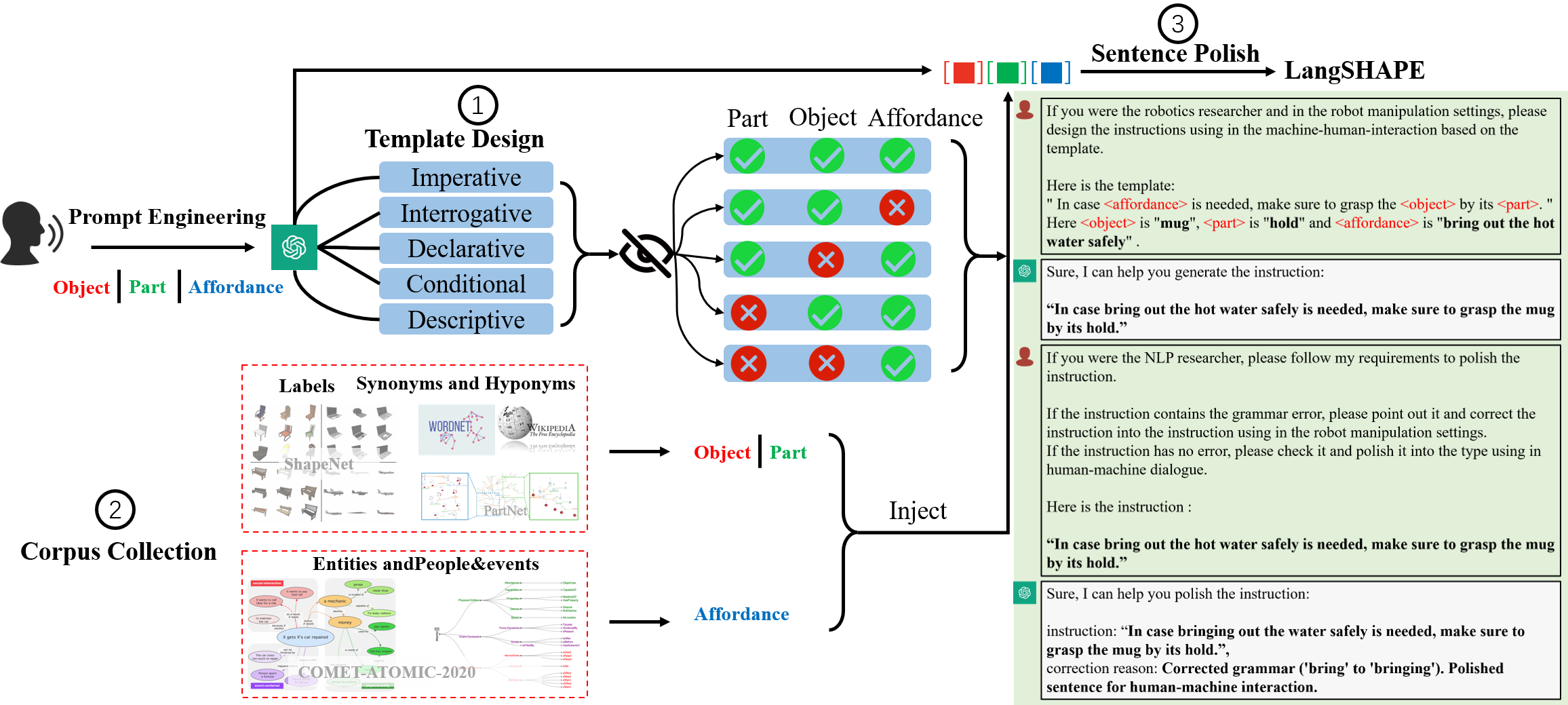}
	\caption{The pipeline of language generation in LangSHAPE. It includes three key steps. The \textbf{first} is to generate various sentence templates by prompt engineering. The \textbf{second} is to collect corpus about \textit{object}, \textit{part}, and \textit{affordance} from the open-world knowledge base. The \textbf{third} is to inject corpus into the template to generate sentences and polish them in expression and grammar.}
	\label{fig:dataset_gen}
\end{figure*}

\section{Problem Statement}
For point cloud based grasp pose detection method~\cite{ten2017grasp}, given a point cloud and a description of the geometry of a robotic hand (hand configuration), \blackcallig{grasp pose detection} is to predict the grasp pose based on the hand configuration, from which a grasp would be formed if the fingers are to close. A typical technical pipeline is to sample enough grasp pose candidates following a designed rule and select the one with the highest score. It can be formulated as a probability model
$P({g_i}|\mathcal{R},\mathbb{C},\Theta)$.
$g_i$ is a sampled grasp pose, $\mathbb{C}$ is a point cloud observation, $\mathcal{R}$ is an interesting region to sample grasp configuration candidates, and $\Theta$ is a hand configuration of a robot.
However, most existing work considers the region of interest (ROI) as prior input from object detection or localization in a scenario (e.g., a cluttered scene). These realize a coarse-grained grasping in object wise, while part semantic of an object is ignored. 

In this paper, we consider \blackcallig{fine-grained grasping} \blackcallig{detection in part wise} by constraining ROI using affordance and intention grounding. We consider a part-aware joint probability model for grasp pose detection using external knowledge in the form of natural language and decompose it into two parts, given by:
\begin{equation}\label{eq:joint_probability}
P({g_i},\mathcal{R}|\mathbb{C},\Theta,\mathcal{Q}) = P(g_i|\mathcal{R},\mathbb{C},\Theta)\times P(\mathcal{R}|\mathbb{C}, \mathcal{Q}),
\end{equation}
where $\mathcal{Q}$ is a natural language sentence which could be a fine-grained object description or human instruction. $P({g_i}|\mathcal{R},\mathbb{C},\Theta)$ is given by a grasp pose detection model (e.g., GPD). $P(\mathcal{R}|\mathbb{C}, \mathcal{Q})$ is given by a 3D part language grounding model. It is noted that natural language in the proposed model is a broad intermediate interface to transfer cognitive knowledge to a robot from human-robot or robot-LLM (large language model, e.g., ChatGPT), robot-structured knowledge base (e.g., knowledge graph), etc. Therefore, we attempt to design a unified method for robot autonomous decision-making and human-in-the-loop decision-making.

The input of the proposed joint probability model consists of a point cloud observation $\mathbb{C}$, natural language sentence $\mathcal{Q}$, and a static hand configuration $\Theta$. The model outputs 3D part grounding prediction in point cloud and a confidence score for a given sampled 6-DoF grasping candidate.
There are two assumptions as follows:

\noindent \textbf{Assumption 1.} High-level semantics from cognition science could provide useful external knowledge (e.g., commonsense) to promote robot decisions positively. We inject this information in the form of natural language.

\noindent \textbf{Assumption 2.} To investigate the relationship between graspability and object-part semantics, we set that there were at least one positive grasping candidate that can be detected within part grounding region.

\section{Data Generation}\label{sec:data_generation}
To model the fine-grained grasping detection in part wise, we collect a dataset named \textbf{LangSHAPE} including point cloud observation \textbf{C}, 6-DoF grasping pose with label $\mathbb{G}$, and natural language $\mathbb{Q}$ about fine-grained grasping. An example is shown in Fig.~\ref{fig:example_langshape}.

\textbf{Firstly}, for \underline{natural language $\mathbb{Q}$}, we choose ShapeNet part dataset~\cite{qi2017pointnet} to obtain the geometric feature and basic semantics about grasped objects and parts, which contains $16,881$ from $16$ categories, annotated with $50$ parts in total. To complete point cloud observation rendering and grasping labeling, we retrieve object meshes from ShapeNetCore~\cite{shapenet2015}. Based on basic semantics from ShapeNet part dataset~\cite{qi2017pointnet}, we augment the corpus using various general knowledge bases, which are used to generate natural language sentences about fine-grained object description or human instruction~\cite{hwang2021comet,mo2019partnet,fellbaum2010wordnet,wikipedia2004wikipedia}. Comparisons with the existing grasping datasets are shown in TABLE~\ref{tab:dataset_compare}.
\textbf{Secondly}, for \underline{point cloud observation \textbf{C}}, deriving from the definition of GPD~\cite{ten2017grasp}, let $\mathcal{W} \subseteq {\mathbb{R}^3}$ denote the robot workspace and $\mathcal{C}_{raw} \subset \mathcal{W}$ the 3D point cloud perceived by the sensor. We extend $\mathcal{C}_{raw}$ to $\mathcal{C} \subseteq \mathbb{R}^4$ with extra part semantic information. \syx{Each point in the point cloud is paired with at least one viewpoint with camera pose $\mathcal{V} \in SE(3)$~\cite{ten2017grasp}. A point cloud observation of an object can be defined as a tuple $ \mathbb{C} = (\mathcal{C}, \mathcal{V}) \in \textbf{C}$.} 
\textbf{Thirdly}, for \underline{grasping label $\mathbb{G}$}, we denote a grasp pose in 3D space $g = (\mathbf{p}, \mathbf{R}) \in SE(3)$, which specifies the position and orientation of the grasping center point of the gripper to the robot base frame. Grasping label paired $label$ with grasping pose is as a $(g, label) \in \mathbb{G}$. One sample in our LangSHAPE dataset is organized as a 3-tuple $(\textbf{C}, \mathbb{G}, \mathbb{Q})$ and detailed as follows.

\begin{figure*}[t]
	\centering 
	\includegraphics[width=0.9\linewidth]{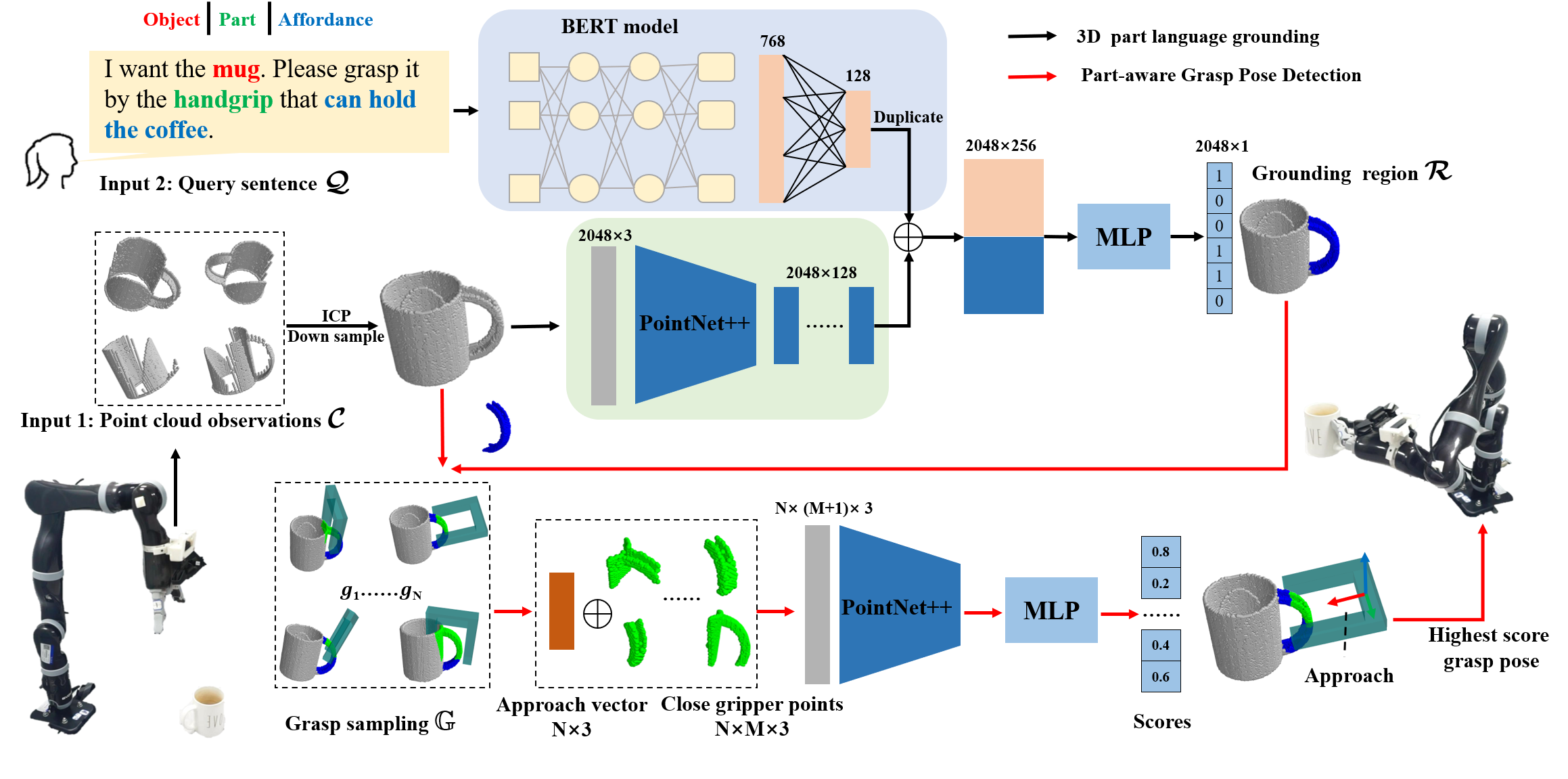}
	\caption{The overall architecture of LangPartGPD. \blackcallig{black} arrow trace refers to 3D part language grounding. The \textcolor{red}{\redcallig{red}} arrow trace refers to part-aware grasp pose detection. Multiple object observations with point cloud are collected, ICP and downsampled before fed into LangPartGPD.}
	\label{fig:network_struture}
\end{figure*}

\subsection{Point Cloud Observation with Semantics}\label{sec:pd_data}
In a real-world setting, a robot can obtain a point cloud observation $\mathbb{C} =(\mathcal{C}, \mathcal{V}) $ to perceive object 3D information using an RGB-D camera. We establish a table scenario to randomly place one object $O$ rendered by \textit{pyrender}\footnote{\url{https://github.com/mmatl/pyrender}} and collect $547,417$ point cloud observations for $42,109$ objects. Specifically, each sample of one object mesh contains $13$ point cloud observations including one omni (full-view) point cloud observation, and $3$ random placement partial-view point cloud observation (each placement includes $4$ viewpoint observations). We adopt part labels in ShapeNet part dataset~\cite{qi2017pointnet} and map them to rendered point cloud by ICP and KD-tree search.

Point cloud via random placement and specific viewpoint can be denoted as $\{\mathbb{C}_{v_i}^{p_j}, i=0,1,2,3; j=0,1,2\}$, in which one point cloud observation from random object placement $p_i$ and four viewpoints $v_i$ is sampled with $4,096$ points $\mathcal{C} \in \mathbb{R}^{4 \times 4096}$ with part label in object coordinates, as shown on the top left of Fig.~\ref{fig:example_langshape}. An omni point cloud $\mathbb{C}_{fullv}$ is sampled from object mesh directly with $10,000$ points $\mathcal{C} \in \mathbb{R}^{4 \times 10000}$  with part label in object coordinates. The whole point cloud observation about one object is denoted as $\textbf{C} =\{\mathbb{C}_{fullv},\mathbb{C}_{v_i}^{p_j}, i=0,1,2,3; j=0,1,2\}$.

\subsection{Grasp Sampling and Labeling}
We follow the sampling policy of PointNetGPD~\cite{liang2019pointnetgpd} using antipodal sampling~\cite{ten2017grasp} based on \textit{trimesh}\footnote{\url{https://github.com/mikedh/trimesh}}. Different from existing work, which samples uniformly on the mesh of the object, we introduce part semantics to the sampling process and only sample feasible grasps on the specific part surface. 
We sample $60$ high-quality grasping candidates from each part surface within an object, which are labeled by executing a grasping operation in our designed MuJoCo~\cite{todorov2012mujoco} simulation environment. Success grasp candidates are labeled as positive grasping labels, while failure grasp candidates are labeled as negative grasping labels. The ratio of the number of success and failure grasping is close to $1:1$ statistically.

Finally, for each sample in our \textbf{LangSHAPE} dataset, we obtain a grasp set $\mathbb{G}= \{(g_{\mathbf{p^i}}^i,label_{\mathbf{p^i}}^i), \mathbf{p^i} \in \mathcal{R}_{part}, i=0,..,59\}$ with $60$ elements, where $g^i$ is the \textit{i}-th grasp pose with position $\mathbf{p^i}$ and orientation $\mathbf{R}^{i}$. $\mathcal{R}_{part}$ is the specific part surface (e.g., surface of \textit{Handle} in a bag, shown on the top right of Fig.~\ref{fig:example_langshape}).

\subsection{Part-affordance-based Language Generation}\label{sec:languae_gen}
The generation of natural language about fine-grained grasping contains three parts shown in Fig.~\ref{fig:dataset_gen}. Firstly, for template design, we adopt ChatGPT~\cite{ChatGPTO40:online}\footnote{Version: GPT-3.5-turbo-0125} to obtain five kinds of language descriptions about object, part, affordance, and human intention, covering \textit{imperative sentence}, \textit{interrogative sentence}, \textit{declarative sentence}, \textit{conditional sentence}, and \textit{descriptive sentence} with $99$ templates. 

Secondly, we collect corpus about object, part, affordance, and human intentions to complete the templates. For object and part, we extract $35$ parts to form part-level labels from ShapeNet part dataset~\cite{qi2017pointnet}. Object and part semantic labels are augmented by considering synonyms and hyponyms in PartNet~\cite{mo2019partnet}, WordNet~\cite{fellbaum2010wordnet}, and Wikipedia~\cite{wikipedia2004wikipedia}. \syx{For affordance and human intention, we collect a mass of scene knowledge from a commonsense knowledge graph COMET-ATOMIC-2020~\cite{hwang2021comet} with $1.33$M daily inferential knowledge tuples about entities and people\&events\footnote{\url{https://mosaickg.apps.allenai.org/model_comet2020_people_events}}.} It represents a large-scale common sense repository of textual descriptions that encode both the social and the physical aspects of common human everyday experiences. We extract corpus based on the tuples and people\&events related to fine-grained grasping and human-robot interaction to complete the above templates. \revision{To guarantee the consistency of grasping operation with affordance, we conduct Query-answer tests using corpus and target parts from our LangSHAPE dataset based on GPT-4\footnote{Version: gpt-4o-2024-08-06}. Existing researches~\cite{liu2023g,klingefjord2024human,norhashim2024measuring,achiam2023gpt} provide evidence that GPT-4 could realize value alignment with humans, therefore we use it to substitute real humans for high-efficient consistency tests.  The consistent rate is $81.5\%$ average with a standard deviation of $0.00754$ by 5 tests, in which $1,000$ samples are sampled randomly each time. Specifically, GPT-4 is asked to play a human to judge given instruction whether indicating the groundtruth target part\footnote{More details are available in our supplementary materials.}.}

Thirdly, the collected corpus is injected into the $99$ templates by in-context learning based on ChatGPT~\cite{ChatGPTO40:online} to generate the final sentence set $\mathbb{Q}$ in \textbf{LangSHAPE}, which includes more than $1.38$ million sentences $\mathcal{Q}$ with corrupted partial information. The generated language sentences are organized into $7$ subsets to test the inference ability of proposed models shown in TABLE~\ref{table:data_org}.

\begin{table*}[t]
\centering
\caption{Multi-level complexity language configuration.}
\small
\scalebox{0.95}{
\begin{tabular}{@{}ll@{}}
\toprule
\multicolumn{1}{c}{\textbf{Langauge Mode}} & \multicolumn{1}{c}{\textbf{Definition}}                                                                    \\ \midrule
full\_data                                 & all $99$ sentences.                                                        \\
known\_all                                 & sentences containing object name, part name, and affordance (intention).      \\
part\_known\_object\_unknown                            & sentences containing part name and affordance, but not containing object name.                               \\
part\_unknown\_object\_known                              & sentences containing object name and affordance, but not containing part name.   \\
part\_unknown\_object\_unknown                 & sentences only containing affordance, not containing either object name or part name. \\
part\_object\_fragment                             & simple fragment of object and part.                        \\
one\_best\_part                             & under human intervention to give an optimal grasp part for each object observation.                        \\ \bottomrule
\end{tabular}
}
\label{table:data_org}
\end{table*}

\section{Fine-grained Grasping Method}
We propose a loosely-coupled symbolic grounding based framework to model Eq.~\ref{eq:joint_probability}, named \textbf{LangPartGPD} (\textbf{Lang}uage-\textbf{Part} \textbf{G}rasp \textbf{P}ose \textbf{D}etection). The overall architecture is shown in Fig.~\ref{fig:network_struture}. It consists of two modules, where the first is a 3D part language grounding model, which is used for $P(\mathcal{R}|\mathbb{C}, \mathcal{Q})$. The second is a part-aware grasp pose detection model for $P({g_i}|\mathcal{R},\mathbb{C},\Theta)$. Since we convert the point cloud from different view into object coordinate representation in Sec.~\ref{sec:pd_data}, we directly use $\mathcal{C}$ to depict point cloud processing in the following descriptions.

\subsection{3D Part Language Grounding}
Given a query sentence $\mathcal{Q} \in \mathbb{Q}$ from human and robotic point cloud observation $\mathcal{C} \in \mathbb{C}$, our 3D language grounding model is to detect the query-related region $\mathcal{R}$ in part wise. To determine if a point in the point cloud belongs to the query region, it can be formulated as a binary classifier function $\phi$ for each point in point cloud $\mathcal{C}$: $(\mathcal{Q}, \mathcal{C}) \to \mathcal{R}\{0,1\}$. 
To achieve this, our proposed model consists of four modules: language encoder, point cloud encoder, multimodal alignment module, and a binary classifier, shown in Fig.~\ref{fig:network_struture} with \blackcallig{black trace}. 

A query sentence $\mathcal{Q}$ from a human is fed to a pre-trained language model encoder (we use BERT\cite{kenton2019bert}\footnote{We use bert-base-uncased model in our work.}) passing two fully connected layers to calculate a 128-dimension language feature $Z_q \in \mathbb{R}^{1 \times 128}$. 
\syx{For point cloud, we use single-view point cloud during model training and simulation process because of the dense rendering point cloud, while we choose four-view point cloud to merge a relatively complete point cloud by iterative closest point (ICP) to camera coordinates in physical experiment because of sparse point cloud sampling from RGB-D camera. $2,048$ points\footnote{We simplify to ignore viewpoint $\mathcal{V}$ representation since we have transformed all points to the same coordinates by ICP.} $\mathcal{C} \in \mathbb{R}^{2048 \times 3}$ are downsampled from the chosen point cloud.} The preprocessed $\mathcal{C}$ is input into PointNet++~\cite{qi2017pointnet++} to encode a feature map $Z_c \in \mathbb{R}^{2048 \times 128}$.
After extract language $Z_q$ and point cloud features $Z_c$, we repeat the $Z_q$ $2,048$ times to construct a feature map $Z'_q \in \mathbb{R}^{2048 \times 128}$. We concatenate $Z_c$ and $Z'_q$ to extract a shared feature map $Z_{fused}$ to realize multimodal alignment using a multilayer perceptron (MLP). \syx{At last, the shared feature is input to a binary classifier to predict which points belong to the grounded region $\mathcal{R}$.} The whole pipeline can be formulated as:

\begin{equation}
\small
\begin{aligned}
{Z_q} &= {E_{lang}}\left( \mathcal{Q} \right),\\
{Z_c} &= {E_{point}}\left( \mathcal{C} \right),\\
{Z_{fused}} &= MLP\left({repeat\left( {{Z_q}} \right) \oplus {Z_c}} \right),\\
\mathcal{R} &= Classifier\left( {{Z_{fused}}} \right),
\end{aligned} 
\end{equation}
where $E_{lang}$ and $E_{point}$ are language and point cloud encoders respectively. $\oplus$ denotes the concatenation operation.

\subsection{Part-aware Grasp Pose Detection}
To achieve part-aware grasp pose detection, we extend PointNetGPD~\cite{liang2019pointnetgpd} in candidate sampling policy and grasp selection. 
Under \textbf{Assumption 2}, different from sampling uniform randomly on the preprocessed point cloud of the whole object \cite{liang2019pointnetgpd}, \syx{we introduce high-level cognitive semantic $\mathcal{R}$ to constrain sampling region for candidate grasp set $g_i \in \mathcal{R}$, which imprecisely represents the position $\mathbf{p^i}$ of grasping $g_i$ close to the region surface $\mathcal{R}$ via the nearest neighbor search, shown in Fig.~\ref{fig:network_struture} with \redcallig{red trace}}. Firstly, grasping pose sampling is taken within region surface $\mathcal{R}$ from 3D language grounding. Then in-hand point cloud is clipped from the whole point cloud and transformed into gripper coordinates concatenated with the approach vector from the grasping pose, which is encoded by PointNet++~\cite{qi2017pointnet++}. Finally, all grasping pose candidates are scored by an MLP and the grasping pose with the highest score is executed by the robot arm. The whole pipeline can be formulated as follows:

\begin{equation}
\small
\begin{aligned}
{\mathbb{G}} &= {Samper_{grasp}}\left( \mathcal{C},\mathcal{R} \right),\\
{\mathcal{C}_{in\_hand}} &= {Filter}\left( \mathcal{C}, g_i \right), g_i \in \mathbb{G}\\
{Z_g} &= {E_{grasp}}\left({g_i.\mathbf{R}.approach \oplus \mathcal{C}_{in\_hand}} \right),\\
{Score} &= MLP\left(Z_g \right),\\
\end{aligned} 
\end{equation}
where $Samper_{grasp}$ is part-aware grasping sampler. $Filter$ is to clip and transform in-hand point cloud into gripper coordinates. $E_{grasp}$ is a grasping feature encoder using PointNet++, in which $g_i.\mathbf{R}.approach \in \mathbb{R}^{1 \times 3}$ is the approach vector of one sampled grasping pose's orientation matrix $\mathbf{R}$ ($g = (\mathbf{p}, \mathbf{R}) \in SE(3)$ denoted in Sec.~\ref{sec:data_generation}).

\subsection{Training and Inference}\label{sec:training_detals}
We train the 3D language grounding model and part-aware grasp pose detection model separately. 
To train 3D language grounding model, we use $(\textbf{C}, \mathbb{Q})$ in our proposed dataset LangSHAPE in Sec.~\ref{sec:data_generation}. The parameters of pre-trained BERT are frozen during the training process.  We use a binary-class cross-entropy loss to optimize the network with Adam optimizer. We train the network for $160$ epoches with batchsize $128$ and learning rate $1e^{-3}$. 
To train the part-aware grasp pose detection model, we use $(\textbf{C}, \mathbb{G})$ in LangSHAPE dataset, in which the oracle point cloud semantic region is used to constrain the sampling region. We also use a binary-class cross-entropy loss to optimize the network with Adam optimizer. The batchsize is $64$, training epoch is $60$, and learning rate is $5e^{-3}$. 

During inference, the language and point cloud input are from LangSHAPE in simulation experiments and from human or LLM and RGB-D camera in real-world physical experiments respectively. The grounding region of an object predicted from 3D language grounding model is fed into the part-aware grasp pose detection model. A series of grasp candidate scores are predicted finally. We select the optimal grasp to execution considering these scores and robotic reachability. All models are trained and tested under PyTorch 1.10 with one Intel Core i9-13900K and two RTX A6000 GPUs.

\begin{figure}[t]
	\centering 
	\includegraphics[width=0.98\linewidth]{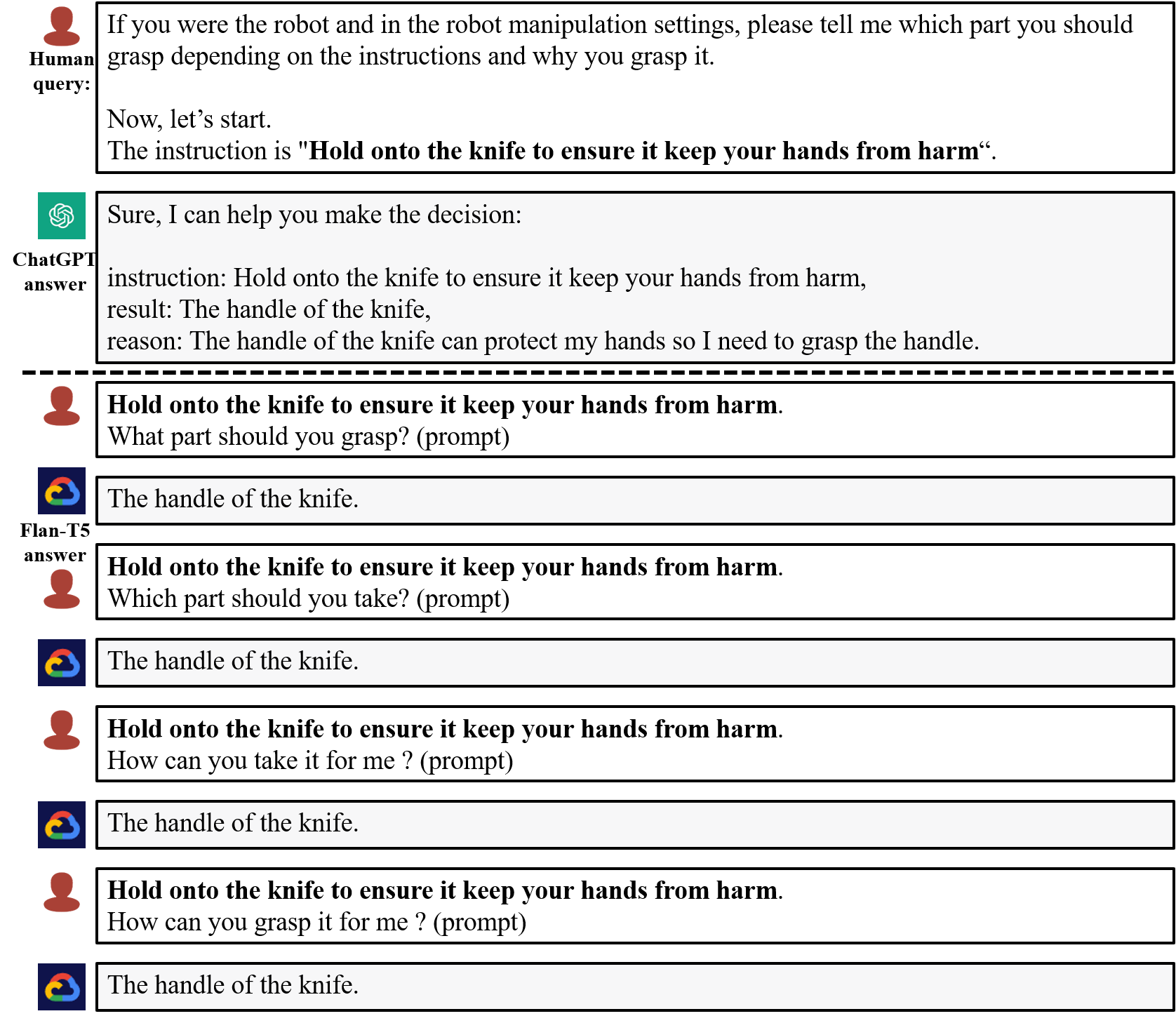}
	\caption{The designed prompts of LangPartGPD-ChatGPT/Flan-T5 to predict grasped part based on corrupted instruction.}
	\label{fig:T5_model}
\end{figure}

\section{Experiment}
We conduct both simulation and real-world physical robot experiments to investigate four research questions (\textbf{RQs}). The first one is regarding the usefulness of our proposed new dataset, the middle two are to analyze the performance of the proposed models for language grounding task and grasping detection task, and the last one is concerning the practicability of our method on a real robot platform.

\noindent \textbf{RQ1:} For the effectiveness of proposed dataset, is our proposed new dataset useful for fine-grained 3D robotic grasping tasks, especially for affordance-aware task?

\noindent \textbf{RQ2:} How does 3D part grounding model using pre-trained language model perform on our proposed LangSHAPE in robustness, generalization, and inference ability?

\noindent \textbf{RQ3:} For fine-grained grasping, how does our proposed part-aware grasping detection predict the 6-DoF grasping pose? What is the superiority of our part-grounding based method? 

\noindent \textbf{RQ4:} For real-world deployment, does our proposed method perform effectively on a real-world robot?

\subsection{Data Organization}\label{sec:data_org}
As introduced in Sec.~\ref{sec:training_detals}, to train and test our proposed models, we split our LangSHAPE dataset based on the inputs of two proposed modules:

\noindent(1) \textbf{Part-aware Grasp Pose Detection:}
\syx{we split our LangSHAPE dataset by object-wise and the part-wise modes, named \textbf{Split Mode}, which is to model grasping detection task on conventional (coarse-grained, only object categories known) split mode and part-aware (fine-grained) split mode respectively.} In detail, For object-wise mode, we split all samples in LangSHAPE by the object category (16 categories) with ratios $(8:1:1)$ for (training/validation/test) sets. Similarly, for part-wise mode, we split all samples in LangSHAPE by the part category (35 categories) with ratios $(8:1:1)$ for (training/validation/test) sets. 

\noindent(2) \textbf{3D Part Language Grounding:} \syx{we set up fine-grained language configurations, named \textbf{Language Mode}, defined in TABLE~\ref{table:data_org}, which is to evaluate the cognitive inference ability of the grounding model across a spectrum of textual complexities.}
We further provide two compositional generalization sets in TABLE~\ref{table:exp3}. Two extra language configurations are introduced:

\begin{itemize}
	\itemsep0em 
\item{\textbf{related data}} has two attributes. First, the object categories of samples in the training set do not occur in the test set. Second, at least one but not all parts of each sample in the training set are similar to those in the test set. Nevertheless, the categories of parts contained in the training set cover the categories of parts in the test set. 
The details are shown in TABLE~\ref{table:exp3} with \textbf{Compositional Factors}. In the first setting, chair, laptop, and skateboard samples in LangSHAPE are collected as the training set, in which they have at least one part such as leg or board. Table samples are used as the test set, which both consists of legs and boards. In the second setting, guitar and pistol samples in LangSHAPE are used as the training set, in which they have at least one part such as body or handle. Mug samples are adopted as the test set, which both consists of body and handle. 

\item{\textbf{non\_related data}} Samples in the training set do not contain any objects or parts that occur in the test set.
\end{itemize}

\begin{table*}[t]
\centering
\caption{Overall results of 3D part language grounding in robustness and generalization.}
\small
\resizebox{0.85\textwidth}{!}{
\begin{tabular}{@{}clccccc@{}}
\toprule
\textbf{Split Mode}          & \multicolumn{1}{c}{\textbf{Model}} & \textbf{Accuracy($\uparrow$)} & \textbf{Part avg IoU($\uparrow$)} & \textbf{Object avg IoU($\uparrow$)} & \textbf{Instance avg IoU($\uparrow$)} & \textbf{Language Encoder} \\ \midrule
\multirow{3}{*}{Part-wise}   & Zero-grounding                     & 0.526             & 0.240                 & 0.275                    & 0.244                     & BERT                      \\
                             & Scratch-grounding                  & 0.896             & 0.556                 & 0.589                    & 0.717                     & Transformer               \\
                             & \textbf{LangPartGPD (ours)}        & \textbf{0.923}    & \textbf{0.640}        & \textbf{0.680}           & \textbf{0.779}            & BERT                      \\ \midrule
\multirow{3}{*}{Object-wise} & Zero-grounding                     & 0.526             & 0.217                 & 0.250                    & 0.224                     & BERT                      \\
                             & Scratch-grounding                  & 0.903             & 0.579                 & 0.608                    & 0.731                     & Transformer               \\
                             & \textbf{LangPartGPD (ours)}        & \textbf{0.920}    & \textbf{0.608}        & \textbf{0.648}           & \textbf{0.776}            & BERT                      \\ \bottomrule
\end{tabular}
}
\label{table:exp1}
\end{table*}

\begin{table*}[t]
\centering
\caption{Comparisons of inference performance with different corrupted languages. For \textbf{Split Mode}, the top half is in part wise. The bottom half is in object wise. For \textbf{Language Encoder}, Scratch-grounding uses Transformer while our LangPartGPD uses BERT.}
\small
\scalebox{0.7}{
\begin{tabular}{@{}llcccccc@{}}
\toprule
\textbf{Split Mode}           & \multicolumn{1}{c}{\textbf{Model}}           & \textbf{Language Mode}         & \textbf{Accuracy($\uparrow$)} & \textbf{Part avg IoU($\uparrow$)} & \textbf{Object avg IoU($\uparrow$)} & \textbf{Instance avg IoU($\uparrow$)} & \textbf{Language Encoder}    \\ \midrule
\multirow{9}{*}{Part-wise}   & \multirow{2}{*}{Scratch-grounding}           & known\_all                     & 0.893             & 0.549                 & 0.608                   & 0.705                     & \multirow{2}{*}{Transformer} \\
                             &                                              & part\_unknown\_object\_known   & 0.897             & 0.551                 & 0.599                   & 0.719                     &                              \\ \cmidrule(lr){2-2} \cmidrule(l){8-8} 
                             & \multirow{5}{*}{\textbf{LangPartGPD (ours)}} & part\_object\_fragment         & \textbf{0.939}    & \textbf{0.664}        & \textbf{0.716}          & \textbf{0.808}            & \multirow{7}{*}{BERT}        \\
                             &                                              & known\_all                     & 0.930             & 0.652                 & 0.694                   & 0.794                     &                              \\
                             &                                              & part\_known\_object\_unknown   & 0.924             & 0.649                 & 0.686                   & 0.774                     &                              \\
                             &                                              & part\_unknown\_object\_known   & 0.913             & 0.634                 & 0.675                   & 0.769                     &                              \\
                             &                                              & part\_unknown\_object\_unknown & 0.918             & 0.652                 & 0.677                   & 0.773                     &                              \\ \cmidrule(lr){2-2}
                             & \textbf{LangPartGPD-Flan-T5 (ours)}          & part\_unknown\_object\_known   & 0.918             & 0.616                 & 0.681                   & 0.782                     &                              \\
                             & \textbf{LangPartGPD-ChatGPT (ours)}          & part\_unknown\_object\_known   & 0.550             & 0.325                 & 0.406                   & 0.335                     &                              \\ \midrule
\multirow{9}{*}{Object-wise} & \multirow{2}{*}{Scratch-grounding}           & known\_all                     & 0.891             & 0.542                 & 0.583                   & 0.707                     & \multirow{2}{*}{Transformer} \\
                             &                                              & part\_unknown\_object\_known   & 0.875             & 0.517                 & 0.574                   & 0.673                     &                              \\ \cmidrule(lr){2-2} \cmidrule(l){8-8} 
                             & \multirow{5}{*}{\textbf{LangPartGPD (ours)}} & part\_object\_fragment         & \textbf{0.934}    & \textbf{0.649}        & \textbf{0.700}          & \textbf{0.805}            & \multirow{7}{*}{BERT}        \\
                             &                                              & known\_all                     & 0.928             & 0.646                 & 0.675                   & 0.793                     &                              \\
                             &                                              & part\_known\_object\_unknown   & 0.926             & 0.639                 & 0.676                   & 0.787                     &                              \\
                             &                                              & part\_unknown\_object\_known   & 0.912             & 0.638                 & 0.672                   & 0.765                     &                              \\
                             &                                              & part\_unknown\_object\_unknown & 0.904             & 0.631                 & 0.661                   & 0.757                     &                              \\ \cmidrule(lr){2-2}
                             & \textbf{LangPartGPD-Flan-T5 (ours)}          & part\_unknown\_object\_known   & 0.913             & 0.605                 & 0.659                   & 0.777                     &                              \\
                             & \textbf{LangPartGPD-ChatGPT (ours)}          & part\_unknown\_object\_known   & 0.587             & 0.320                 & 0.351                   & 0.291                     &                              \\ \bottomrule
\end{tabular}
}
\label{table:exp2}
\end{table*}

\subsection{Evaluation Metrics}
Grounding evaluation and 6-DoF grasping detection evaluation are defined as follows. 

\noindent For 3D part language grounding, we use four metrics derived from 3D part segmentation~\cite{qi2017pointnet,qi2017pointnet++}:

{\textbf{$\bullet$ Accuracy:}} Since we formulate 3D part language grounding as a binary classification problem, we calculate classification accuracy on points.

{\textbf{$\bullet$ Part avg IoU:}} From the part-level perspective, we calculate the IoU of grounded points for each sample~\cite{qi2017pointnet} and average IoUs based on part class to calculate mIoUs. Finally, we average each kind of part's mIoUs to calculate the \textit{Part avg IoU}.

{\textbf{$\bullet$ Object avg IoU:}} From the object-level perspective, we calculate the IoU of grounded points in each sample and average IoUs based on object class to get mIoUs. Finally, we average each kind of object's mIoUs to calculate the \textit{Object avg IoU}.

{\textbf{$\bullet$ Instance avg IoU:}} Generally, we calculate the IoU of grounded points in each sample and average all IoUs directly, which is used to select models during the training process. 

\noindent For 6-DoF grasping detection, we give \syx{four} metrics:

{\textbf{$\bullet$ Part-specific Success Rate:}} The percentage of grasps that require both grasp points grounding to be correct and predicted grasping execution to be successful.

{\textbf{$\bullet$ Part-agnostic Success Rate:}} The percentage of grasps that only require the predicted grasping execution to be successful.

\syx{{\textbf{$\bullet$ Relative Increase:}} It indicates a relative improvement in success rate between random sampling or GPD-based grasping-selected policy within antipodal high-quality grasping candidates~\cite{ten2017grasp}, which is to study the influence between different data \textbf{Split Mode} (i.e. part-wise mode vs. object-wise mode).} for fine-grained grasping detection.

{\textbf{$\bullet$ Trial Cost:}} To get a high-quality grasp candidate for the grasp pose detection, how many grasp sampling trials are needed to perform using the antipodal grasping sampler (GPG)~\cite{ten2017grasp}.

\subsection{Model Configurations}
\noindent We design five baselines for comparisons. The first two are about 3D part language grounding, which consider the zero-shot method and model trained from the scratch method. The latter two are about 6-DoF grasping detection, which consider sampling region constraint and selection of grasping from candidates.

{\textbf{$\bullet$ Zero-grounding:}} For 3D part language grounding, inspired by \cite{thomason2022language}, we compare our method with a zero-shot classifier using pre-trained models directly. Instead of finetuning BERT, we use cosine distance between visual and language features to predict whether each point is grounded or not. Visual encoder is adopted from a pre-trained part segmentation model\cite{qi2017pointnet++}, while language encoder is BERT with frozen parameters.

{\textbf{$\bullet$ Scratch-grounding:}} For 3D part language grounding, we replace BERT in our proposed method in Fig.~\ref{fig:network_struture} with a Transformer encoder\footnote{\url{https://github.com/pytorch/examples/tree/master/word_language_model} (\textbf{6 encoder\_layers implemented})}, and train the whole model from scratch. This is to verify whether the pre-trained model can provide useful prior knowledge for 3D part understanding.

{\textbf{$\bullet$ PointNetGPD:}} For 3D grasp pose detection, we use PointNetGPD \cite{liang2019pointnetgpd} without 3D part language grounding intervention during the sampling process as a baseline to verify the superiority of the language-guided method.

{\textbf{$\bullet$ Random-grasp:}} For 3D grasp pose detection, random-sampled grasping from grasping candidates generated from GPD method~\cite{ten2017grasp} is used to execute grasping operation without scores predicted by PointNetGPD.

{\textbf{$\bullet$ Contact-GraspNet:}} \revision{is an end-to-end network to generate 6-DoF parallel-jaw grasps from a depth image directly~\cite{sundermeyer2021contact}. For 3D grasp pose detection, we set up three kinds of models and variants: (1) naive Contact-GraspNet; (2) Contact-GraspNet with our proposed 3D Part Language Grounding to realize part-aware grasping; (3) Contact-GraspNet with FastSAM~\cite{zhao2023fast} to realize part-aware grasping.}

\noindent We subdivide our proposed method into two kinds of models for 3D part language grounding and grasp pose detection:

{\textbf{$\bullet$ LangPartGPD}} is what we propose in Fig.~\ref{fig:network_struture}.

{\textbf{$\bullet$ LangPartGPD-\{\textit{LLM}\}:}} To empower the bidirectional ability of human-robot interaction, we introduce an extra large language model (LLM) (ChatGPT~\cite{ChatGPTO40:online}\footnote{Version: GPT-3.5-turbo-0125} and Flan-T5~\cite{chung2022scaling}) to infer 3D part language grounding based on designed prompts in zero-shot and finetuning setting respectively. For \underline{LangPartGPD-Flan-T5}, the instruction from human is first fed into Flan-T5 finetuned by prompt learning, in which the input is the language sentence with \blackcallig{part\_unknown\_object\_known}\footnote{Since mentioned models are both unimodal models without visual input, it is necessary to inform object name to assist grounding part inference for this kind of models.}.
The output of \underline{LangPartGPD-Flan-T5} and \underline{LangPartGPD-ChatGPT} are both \blackcallig{part\_object\_fragment} covering object and part in Sec.~\ref{sec:languae_gen}. The pipeline is shown in Fig.~\ref{fig:T5_model}.

\begin{table*}[t]
\centering
\caption{Results of compositional generalization with two subsets in LangSHAPE dataset.}
\small
\scalebox{0.73}{
\begin{tabular}{@{}llccccccc@{}}
\toprule
\multicolumn{1}{c}{\textbf{Model}}       & \multicolumn{1}{c}{\textbf{Language Mode}} & \textbf{Accuracy($\uparrow$)} & \textbf{Part avg IoU($\uparrow$)} & \textbf{Part\_1 IoU($\uparrow$)}                                      & \textbf{Part\_2 IoU($\uparrow$)}                                        & \textbf{Instance avg IoU($\uparrow$)} & \textbf{Training Samples} & \textbf{Test Samples}                                            \\ \midrule
\multicolumn{2}{c}{\textbf{Compositional Factor:}}                                    &                   &                       & \begin{tabular}[c]{@{}c@{}}chair-\redcallig{leg}\\ \includegraphics[width=0.5in]{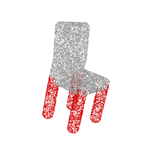}\end{tabular}   & \begin{tabular}[c]{@{}c@{}}skateboard-\bluecallig{board}\\ \includegraphics[width=0.5in]{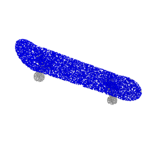}\end{tabular}  & \multicolumn{2}{c}{$\Longrightarrow$}                 & \begin{tabular}[c]{@{}c@{}}table-(\redcallig{leg}, \bluecallig{board})\\ \includegraphics[width=0.5in]{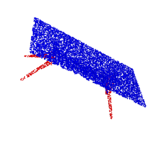}\end{tabular} \\ \cmidrule(l){3-9} 
Zero-grounding                               & related data                               & 0.500             & 0.256                 & 0.186                                                     & 0.326                                                       & 0.257                     & -                         & 10,357                                                           \\ \cmidrule(r){1-1}
\multirow{2}{*}{\textbf{LangPartGPD (ours)}} & related data                               & \textbf{0.715}    & \textbf{0.522}        & \textbf{0.467}                                            & \textbf{0.576}                                              & \textbf{0.522}            & 12,439                    & 10,357                                                           \\
                                         & non\_related data                          & 0.517             & 0.281                 & 0.156                                                     & 0.407                                                       & 0.281                     & 12,568                    & 10,357                                                           \\ \midrule
\multicolumn{2}{c}{\textbf{Compositional Factor:}}                                    &                   & \multicolumn{1}{l}{}  & \begin{tabular}[c]{@{}c@{}}guitar-\redcallig{body}\\ \includegraphics[width=0.5in]{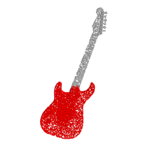}\end{tabular} & \begin{tabular}[c]{@{}c@{}}pistol-\bluecallig{handle}\\ \includegraphics[width=0.5in]{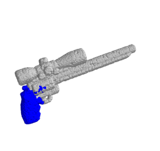}\end{tabular} & \multicolumn{2}{c}{$\Longrightarrow$}                 & \begin{tabular}[c]{@{}c@{}}mug-(\redcallig{body}, \bluecallig{handle})\\ \includegraphics[width=0.5in]{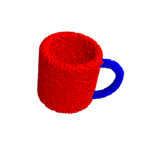}\end{tabular} \\ \cmidrule(l){3-9} 
Zero-grounding                               & related data                               & 0.508             & 0.246                 & 0.432                                                     & 0.059                                                       & 0.247                     & -                         & 350                                                              \\ \cmidrule(r){1-1}
\multirow{2}{*}{\textbf{LangPartGPD (ours)}} & related data                               & \textbf{0.606}    & \textbf{0.472}        & \textbf{0.817}                                            & \textbf{0.127}                                              & \textbf{0.474}            & 2,882                     & 350                                                              \\
                                         & non\_related data                          & 0.482             & 0.257                 & 0.464                                                     & 0.049                                                       & 0.258                     & 12,439                    & 350                                                              \\ \bottomrule
\end{tabular}
}
\label{table:exp3}
\end{table*}

\begin{figure}[t]
	\centering 
	\includegraphics[width=0.95\linewidth]{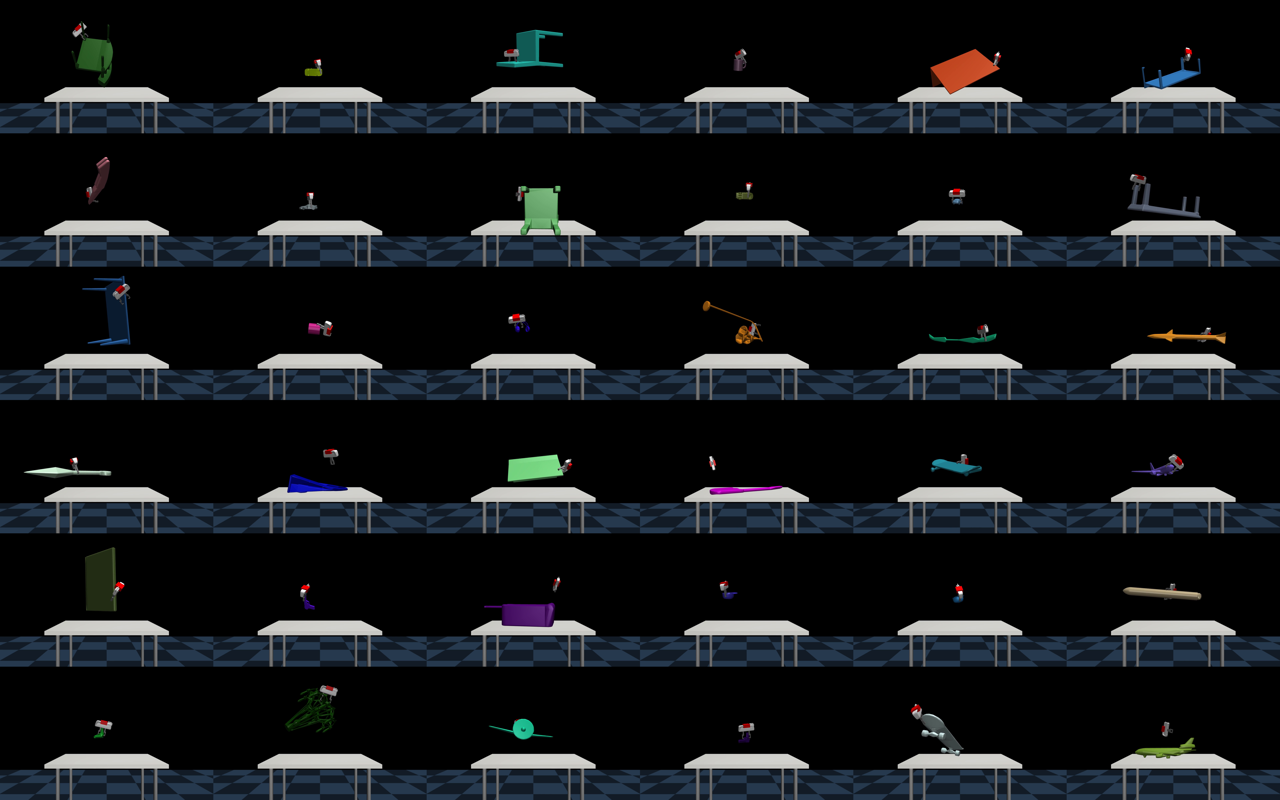}
	\caption{6-DoF Fine-grained grasping experiments in simulation by MuJoCo.}
	\label{fig:sim_exp}
\end{figure}

\subsection{Simulation Experiments on LangSHAPE}
Based on our proposed LangSHAPE dataset, we give a series of quantitative evaluations to answer research questions \textbf{RQ1-RQ3}. The selected models are followed by the maximum Instance avg IoU in the validation set.

\subsubsection{3D Part Language Grounding}
To evaluate the overall grounding performance of our proposed model, we compare our model with Zero-grounding and Scratch-grounding in part wise and object-wise data split mode, shown in TABLE~\ref{table:exp1}. The language mode used in model training is full\_data. For \textbf{RQ1}, our proposed model LangPartGPD outperforms in all metrics. Compared with Zero-grounding ($0.217$ in Part avg IoU), our model ($0.608$ in Part avg IoU) achieves more than $1.8$ times improvement relative to the zero-shot method in object wise. We attribute the poor performance by Zero-grounding to two points. Firstly, there is no domain adaptation to relieve the domain gap between pre-trained domain and LangSHAPE domain. Furthermore, from the feature representation, shared feature space of language and point cloud is missing resulting in poor prediction. Both of these also reflect the necessity of our language-pointcloud-grasp dataset (i.e. LangSHAPE) to realize joint training for fine-grained language-pointcloud joint representation at the object-part level.
For \textbf{RQ2}, compared to Scratch-grounding, our model achieves more than $6\%$ improvement in Instance avg IoU with part wise. The results indicate the advantage of pre-trained language model used in our method over the randomly initialized model (i.e. Transformer) in robustness and generalization. \syx{Notably, from the \textbf{Split Mode}, model trained on part-wise mode achieves better performance in all metrics, which indicates part-wise data split more fit part-level grounding training compared to conventional data split based on object semantics.}

\begin{table*}[t]
\centering

\caption{Comparisons of fine-grained grasping detection in simulation.}
\scalebox{0.8}{
\begin{tabular}{@{}cccccccccc@{}}
\toprule
\textbf{Split Mode}          & \textbf{Model}               & \textbf{\begin{tabular}[c]{@{}c@{}}Grasp Sampling \\ Area\end{tabular}} & \textbf{\begin{tabular}[c]{@{}c@{}}Grasp Candidate \\ Selection\end{tabular}} & \textbf{One Best Part} & \textbf{\begin{tabular}[c]{@{}c@{}}Part-agnostic \\ Success Rate $\uparrow$\end{tabular}} & \textbf{\begin{tabular}[c]{@{}c@{}}Relative Increase\\ $|\vartriangle|\uparrow$\end{tabular}} & \textbf{\begin{tabular}[c]{@{}c@{}}Part-specific \\ Success Rate $\uparrow$\end{tabular}} & \textbf{\begin{tabular}[c]{@{}c@{}}Relative Increase\\ $|\vartriangle|\uparrow$\end{tabular}} & \textbf{Trial Cost} $\downarrow$ \\ \midrule
\multirow{7}{*}{Part-wise}   & \multirow{4}{*}{\textbf{LangPartGPD}} & \multirow{4}{*}{part grounding}                                         & highest-score                                                                 & no                     & 0.819                                                                          & \multirow{2}{*}{0.186}                                          & 0.685                                                                          & \multirow{2}{*}{0.151}                                          & -                   \\
                             &                              &                                                                         & random                                                                        & no                     & 0.633                                                                          &                                                                 & 0.534                                                                          &                                                                 & -                   \\ \cmidrule(lr){6-9}
                             &                              &                                                                         & random                                                                        & yes                    & 0.648                                                                          & \multirow{2}{*}{0.203}                                          & 0.540                                                                          & \multirow{2}{*}{0.165}                                          & -                   \\
                             &                              &                                                                         & highest-score                                                                 & yes                    & \textbf{0.851}                                                                 &                                                                 & \textbf{0.706}                                                                 &                                                                 & \textbf{28.520}     \\ \cmidrule(l){2-10} 
                             & PointNetGPD                  & \multirow{3}{*}{global}                                                 & highest-score                                                                 & no                     & 0.827                                                                          &                                                                 & 0.341                                                                          &                                                                 & -                   \\
                             & Random-grasp                 &                                                                         & random                                                                        & no                     & 0.634                                                                          &                                                                 & 0.254                                                                          &                                                                 & -                   \\
                             & PointNetGPD                  &                                                                         & highest-score                                                                 & yes                    & 0.846                                                                          &                                                                 & 0.398                                                                          &                                                                 & \textbf{33.799}     \\ \midrule
\multirow{7}{*}{Object-wise} & \multirow{4}{*}{\textbf{LangPartGPD}} & \multirow{4}{*}{part grounding}                                         & highest-score                                                                 & no                     & 0.837                                                                          & \multirow{2}{*}{0.177}                                          & 0.701                                                                          & \multirow{2}{*}{0.146}                                          & -                   \\
                             &                              &                                                                         & random                                                                        & no                     & 0.660                                                                          &                                                                 & 0.555                                                                          &                                                                 & -                   \\ \cmidrule(lr){6-9}
                             &                              &                                                                         & random                                                                        & yes                    & 0.665                                                                          & \multirow{2}{*}{0.194}                                          & 0.575                                                                          & \multirow{2}{*}{0.161}                                          & -                   \\
                             &                              &                                                                         & highest-score                                                                 & yes                    & \textbf{0.859}                                                                 &                                                                 & \textbf{0.737}                                                                 &                                                                 & 30.136              \\ \cmidrule(l){2-10} 
                             & PointNetGPD                  & \multirow{3}{*}{global}                                                 & highest-score                                                                 & no                     & 0.842                                                                          &                                                                 & 0.338                                                                          &                                                                 & -                   \\
                             & Random-grasp                 &                                                                         & random                                                                        & no                     & 0.647                                                                          &                                                                 & 0.265                                                                          &                                                                 & -                   \\
                             & PointNetGPD                  &                                                                         & highest-score                                                                 & yes                    & 0.846                                                                          &                                                                 & 0.428                                                                          &                                                                 & 35.990              \\ \bottomrule
\end{tabular}
}
\label{table:exp4}
\end{table*}

\subsubsection{Affordance Inference}
To evaluate the inference ability of the proposed 3D part grounding model, we set up different corrupted language inputs to train models. Four models are used for comparisons: Scratch-grounding, LangPartGPD, LangPartGPD-Flan-T5, and LangPartGPD-ChatGPT. The first two are trained following language mode and split mode in TABLE~\ref{table:exp2}. In LangPartGPD-Flan-T5, we introduce a finetuned Flan-T5~\cite{chung2022scaling} with prompt engineering. We design four prompts similar to~\cite{ahn2022can} shown in Fig.~\ref{fig:T5_model}, one of which is randomly selected. The generative sentence from Flan-T5 or ChatGPT is fed into our LangPartGPD trained with language mode \blackcallig{part\_object\_fragment}.

For \textbf{RQ2} of inference ability, in LangPartGPD, as we can see \blackcallig{known\_all}, \blackcallig{part\_known\_object\_unknown}, and \blackcallig{part\_unknown\_object\_known} models, with different object attributes being corrupted in TABLE~\ref{table:exp2}, the performance of models decreases obviously. 
Generally, benefit from encoding textual information using pre-trained model, our proposed method shows better inference ability using the semantics about objects, parts, and affordance 
than models trained from scratch in know\_all (language mode). For example, in part-wise, compared to Scratch-grounding ($0.549$), our LangPartGPD achieves $0.652$ in Part avg IoU.

We also conduct an ablation study within LangPartGPD with different language modes in TABLE~\ref{table:exp2}. Models trained on \blackcallig{part\_object\_fragment} perform best both in part-wise and object-wise settings, of which models only link symbols of part and object to specific points without reasoning. For the other four language modes, with different numbers of elements corrupted in the sentence, the level of difficulty in reasoning is also continuously increasing, in which for most metrics, the performance is on a downward trend shown in the middle part of part-wise and object-wise results. 
From Accuracy and Instance avg IoU, we can find that when the object name is unknown, the models perform better than the part-unknown conditions generally, which indicates local (part-level) semantics are harder to infer than global semantics (object-level) in fine-grained grounding inference tasks. For example, the model trained on \blackcallig{part\_known\_object\_unknown} achieves $0.924$ while the model trained on \blackcallig{part\_unknown\_object\_known} achieves $0.913$ in Accuracy with part-wise setting.

For \blackcallig{part\_unknown\_object\_known} results in TABLE~\ref{table:exp2}, we can further find that it is necessary to finetune or train from scratch a model for fine-grained 3D part language grounding suffering from insufficiency of language-geometry information alignment in current LLMs. For example, comparing LangPartGPD and LangPartGPD-Flan-T5 with LangPartGPD-ChatGPT (zero-shot method), the former two achieve close good performance with $0.638$ and $0.605$ while the latter is $0.320$ in Part avg IoU (object-wise split mode). This indicates current LLMs still face challenges in handling fine-grained 3D understanding and inference in object-part wise although LLMs have shown impressive performance in many multimodal tasks. Our methods give an effective exploration and achieve competitive performance in 3D part language inference.

\begin{table*}[t]
\centering

\caption{\revision{Comparisons of fine-grained grasping detection on physical robot.}}
\resizebox{0.95\textwidth}{!}{
\begin{tabular}{@{}cl|cccccc@{}}
\toprule
No. & \multicolumn{1}{c|}{Model}      & \multicolumn{2}{c}{known\_all}                                                                                                                & \multicolumn{2}{c}{part\_unknown}                                                                                                             & \multicolumn{2}{c}{object\_part\_fragment}                                                                                                     \\ \cmidrule(l){3-8} 
    & \multicolumn{1}{c|}{}           & \begin{tabular}[c]{@{}c@{}}Part-agnostic \\ Success Rate $\uparrow$\end{tabular} & \begin{tabular}[c]{@{}c@{}}Part-specific \\ Success Rate $\uparrow$\end{tabular} & \begin{tabular}[c]{@{}c@{}}Part-agnostic \\ Success Rate $\uparrow$\end{tabular} & \begin{tabular}[c]{@{}c@{}}Part-specific \\ Success Rate $\uparrow$\end{tabular} & \begin{tabular}[c]{@{}c@{}}Part-agnostic \\ Success Rate $\uparrow$\end{tabular} & \begin{tabular}[c]{@{}c@{}}Part-specific \\ Success Rate $\uparrow$\end{tabular} \\ \midrule
1   & Contact-GraspNet~\cite{sundermeyer2021contact}                & 0.657                                                                 & \multicolumn{1}{c|}{-}                                                & -                                                                     & \multicolumn{1}{c|}{-}                                                & -                                                                     & -                                                                     \\
2   & Contact-GraspNet+part grounding & 0.775                                                                 & \multicolumn{1}{c|}{0.713}                                            & 0.727                                                                 & \multicolumn{1}{c|}{0.675}                                            & 0.649                                                                 & 0.635                                                                 \\
3   & Contact-GraspNet+FastSAM~\cite{zhao2023fast}        & 0.663                                                                 & \multicolumn{1}{c|}{0.388}                                            & 0.662                                                                 & \multicolumn{1}{c|}{0.403}                                            & 0.608                                                                 & 0.378                                                                 \\ \midrule
4   & Random-grasp~\cite{ten2017grasp}                    & 0.509                                                                 & \multicolumn{1}{c|}{-}                                                & -                                                                     & \multicolumn{1}{c|}{-}                                                & -                                                                     & -                                                                     \\
5   & Random-grasp+part grounding     & 0.888                                                                 & \multicolumn{1}{c|}{0.813}                                            & 0.792                                                                 & \multicolumn{1}{c|}{0.766}                                            & 0.770                                                                 & 0.757                                                                 \\ \midrule
7   & PointNetGPD~\cite{liang2019pointnetgpd}                     & 0.704                                                                 & \multicolumn{1}{c|}{-}                                                & -                                                                     & \multicolumn{1}{c|}{-}                                                & -                                                                     & -                                                                     \\
8   & \textbf{LangPartGPD(ours)}               & 0.913                                                                 & \multicolumn{1}{c|}{0.813}                                            & 0.818                                                                 & \multicolumn{1}{c|}{0.779}                                            & 0.784                                                                 & 0.770                                                                 \\ \bottomrule
\end{tabular}
}
\label{table:phy_exp}
\end{table*}

\subsubsection{Compsitional Generalization}
Results in TABLE~\ref{table:exp1} and TABLE~\ref{table:exp2} indicate that within most metrics, models trained on part-wise split mode achieve better performance, especially in Part avg IoU, which seems to indicate that the sample split based on part semantic prefers to part grounding modeling.
To further validate this hypothesis, we analyze and evaluate the performance in compositional generalization.
It is the ability to generalize systematically to a new data distribution by combining known components~\cite{kim2021improving}. 
To measure the compositional generalization of our models, we propose two compositional generalization subsets defined in Sec.~\ref{sec:data_org}, shown in TABLE~\ref{table:exp3}.
For \textbf{RQ2} of generalization, in LangPartGPD, the model trained on related data achieves better performance in all metrics compared with the model trained using non\_related data. For example, in the first subset (the top part of TABLE~\ref{table:exp3}), our method trained on related data outperforms $85\%$ over the model trained on non\_related data ($0.281 \rightarrow 0.522$). This indicates that our proposed model performs effective generalization, even with a few shot samples. Furthermore, data organization considering part-level semantics can promote fine-grained geometric feature understanding during model training. Likewise, compared with Zero-grounding, results show that our proposed model also performs better than the zero-shot method.

\begin{figure}[t]
	\centering 
	\includegraphics[width=0.7\linewidth]{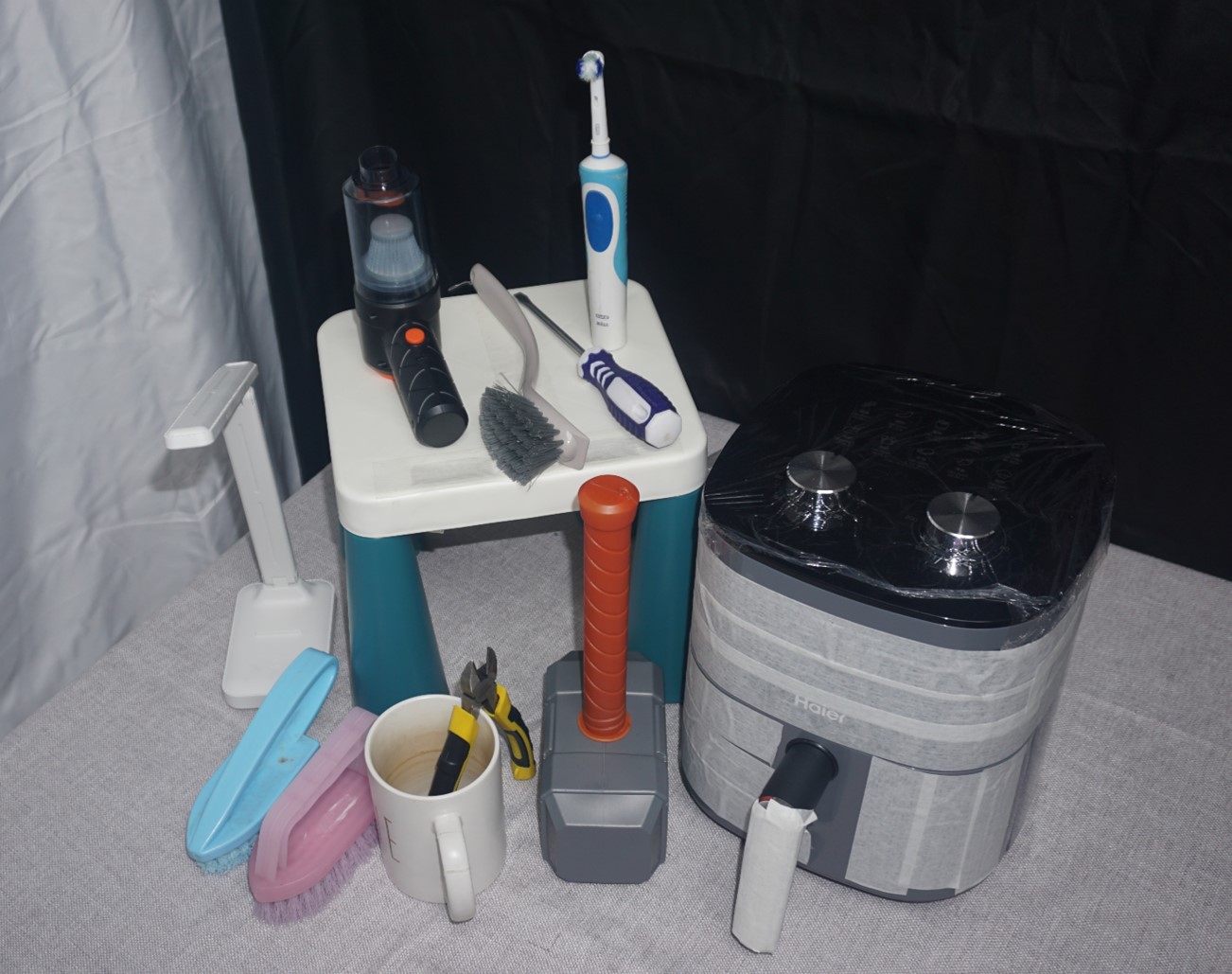}
	\caption{\revision{Novel objects tested in physical robot experiments.}}
	\label{fig:exp_obj}
\end{figure}

\begin{figure*}[t]
	\centering 
	\includegraphics[width=0.7\linewidth]{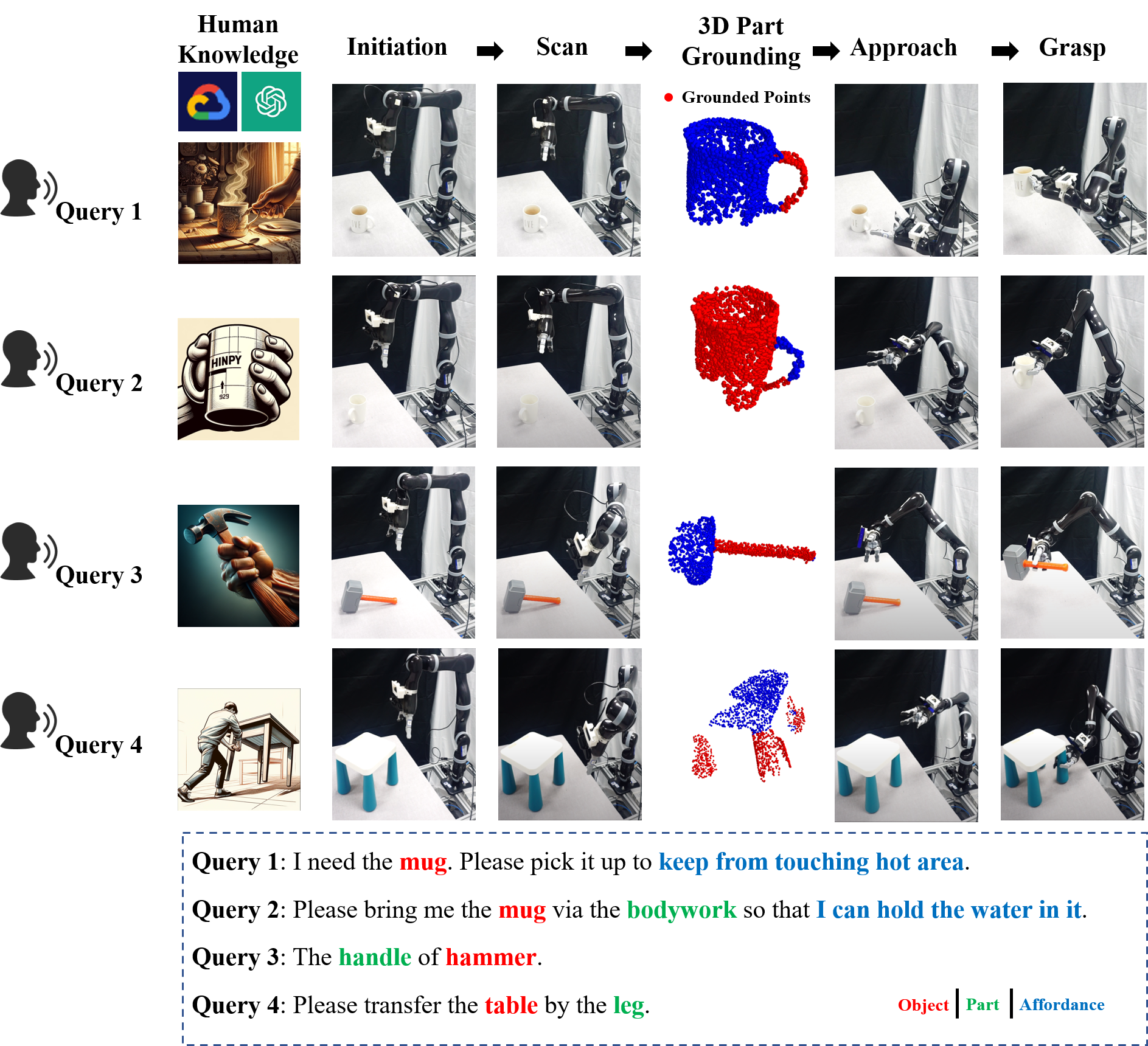}
	\caption{Physical robot grasping pipeline with language query.}
	\label{fig:exp_rob}
\end{figure*}

\subsubsection{Grasping Detection and Cost}
To evaluate the effectiveness of our proposed method in fine-grained 6-DoF grasping detection, we evaluate LangPartGPD on the whole LangSHAPE dataset (including 3D part language grounding and part-aware grasping detection) in simulation (shown in Fig.~\ref{fig:sim_exp}). We train 3D part language grounding and PointNetGPD respectively following the part-wise and object-wise splits respectively. Language mode is part\_object\_fragment in most evaluations. The results are shown in TABLE~\ref{table:exp4}. The grasp sampling rule is that the sampler ends sampling at a maximum of $150$ sample trials or gets $20$ high-quality candidate grasps. 

For \textbf{RQ3}, compared to PointNetGPD (global+highest-score), our LangPartGPD (part\_grounding+highest-score) achieves the close performance of part-agnostic success rate, which is due to the improvements of our method in the sampling process rather than score modeling. In contrast, our method outperforms over $34\%$ of part-specific success rate in both part-wise and object-wise test settings in TABLE~\ref{table:exp4}. 
Since not each part of an object is equally easy to grasp, we propose an extra language mode named one\_best\_part in TABLE~\ref{table:data_org}, in which one best part of each object is specific by human knowledge.
Compared to PointNetGPD (global+highest-score), our LangPartGPD (part\_grounding+highest-score+one\_best\_part) achieves $2.4\%$ and $36.5\%$ improvement of part-agnostic success rate and part-specific success rate in part wise. Even compared to Random-grasp (global+random), our LangPartGPD (part\_grounding+random+one\_best\_part) also achieves similar improvement ($0.648$ vs. $0.634$, $0.540$ vs. $0.254$) in part wise. Furthermore, since the best part is selected to realize part grounding, sampling cost during grasping detection on the grounded region both in part wise and object wise reduce more than $18\%$ relatively ($33.799 \rightarrow 28.520$, $35.990 \rightarrow 30.136$)\footnote{Since we assume our proposed 3D part language grounding is as an interface for humans or LLMs to inject symbolic knowledge assisting in finding an optimized grasping pose, we only consider sampling cost based on the optimized grasping part obtained from external symbolic knowledge.}. From the data split mode, as we can see although the absolute performances of the model trained on part-wise split are weaker than those on object-wise split, Relative Increase results show that models trained on part-wise split possess evident superiority compared to models trained on object-wise data split. By holistic analyzing TABLE~\ref{table:exp2} and TABLE~\ref{table:exp4}, we attribute the absolute performance superiority of models trained on object-wise split to the difference of test set between part-wise and object-wise data split. The antipodal sampler is sensitive to the sampling surface of the grasped object, therefore we propose a relative metric to analyze the performances with different split modes.

\subsection{Physical Robot Experiments}
To evaluate the effectiveness of our proposed method in the real world, we deploy our models on a real robot system to realize part-aware grasping following human instruction by 3D part language grounding. 
We choose a single-arm robot Kinova Jaco 7DoF with three fingers to perform manipulation. An eye-in-hand camera Intel RealSense SR300 and D435 is fixed on the wrist of the end-effector. The system is deployed on a PC running Ubuntu 18.04 and ROS Melodic with one Intel Core i9-13900K and two RTX A6000 GPUs. The intrinsic and extrinsic parameters of the camera are calibrated. We select our LangPartGPD model under the training configuration of part-wise (\textbf{Split Mode}) and full\_data (\textbf{Language Mode}). 
Since our proposed method is to operate in part wise, it requires a more fine-grained perception of the target object. For point cloud collection, we design a multi-view (three) policy to collect each viewpoint cloud and transform it into the robot base frame. All viewpoint point clouds are merged by ICP to get a relatively complete representation of the target object.

\revision{For \textbf{RQ4}, as shown in Fig.~\ref{fig:exp_obj}, we select 12 categories of household objects (most of them are unseen in our proposed LangSHAPE dataset) for quantitative evaluations shown in Fig.~\ref{fig:exp_rob}. Three kinds of grasping models are tested with more than $1,200$ grasping trials, including end-to-end Contact-GraspNet~\cite{sundermeyer2021contact}, GPD~\cite{ten2017grasp}, and PointNetGPD~\cite{liang2019pointnetgpd}. Furthermore, three instruction settings are designed to evaluate the models' fine-grained grasping detection and part inference ability\footnote{\textbf{part\_unknown} is a merged language mode covering part\_unknown\_object\_known and part\_unknown\_object\_unknown modes.}.

Experimental results are reported in TABLE~\ref{table:phy_exp}. It is noted that models No.1,4,7, do not need text inputs, we report their performances in \textbf{known\_all} setting simply. For model No.3 with FastSAM~\cite{zhao2023fast}, since it only
performs well when using words or noun phrases as input, we use part words as input to guarantee FastSAM's performance for the three instruction settings. Specifically, when using FastSAM to ground `handle', we input the text prompt `handle' instead of instructions designed in TABLE~\ref{table:data_org}. \textbf{For RQ3} of general grasping performance, by analyzing models No.1 vs. No.2 ($0.657 \rightarrow 0.775$), No.4. vs. No.5 ($0.509 \rightarrow 0.888$), and No.7 vs. No.8 ($0.704 \rightarrow 0.913$), models intervened by our 3D part grounding achieve at least $11.8\%$ improvements in part-agnostic success rate compared to methods sampling grasping candidates over the global pointcloud observation, which indicates the general advance of region constraint (part-aware) for grasping detection qualities. \textbf{For RQ3} of fine-grained grasping performance, under \textbf{known\_all}, \textbf{part\_unknown} and \textbf{object\_part\_fragment} settings, in the real-world scenario, our proposed 3D part grounding models can ground target regions assisting part-specific grasping pose generation. Performances of models No.2, No.5, and No.8 indicate that more information related to grasping is given, and there is a higher part-specific grasping success rate. All three models achieve more than $67.5\%$ part-specific grasping success rate in \textbf{part\_unknown} instruction settings indicating the 3D part inference ability of our grounding model indirectly in the real world. Furthermore, we also compare the method transferring fine-grained semantics from image space using FastSAM~\cite{zhao2023fast}. Compared to models No.2 and No.3, although FastSAM achieves perfect performance in image segmentation which is mapped into the depth image space by aligning the coordinates of RGB image and depth image, it cannot have part inference ability while only grounding noun phrases to pixels. As above mentioned, we use part words or noun phrases to realize the FastSAM-based model, while model No.2 using our 3D part grounding can still achieve more than $25.7\%$ ($0.388 \rightarrow 0.713, 0.403 \rightarrow 0.675, 0.378 \rightarrow 0.635$) part-specific grasping success rate in three instruction settings. More experiments are available in the attached video\footnote{ \url{https://sites.google.com/view/lang-shape}}.
}

\section{Conclusion} 
We investigate part-level affordance on fine-grained robotic grasping using explicit language as an interface to provide explainable decisions. The LangSHAPE dataset is constructed to facilitate the investigation of part-level robotic cognition. 3D part language grounding and part-aware grasp pose detection model are proposed to allow fine-grained robotic grasping. Benchmark experiments show that our proposed method outperforms 3D part grasp grounding in inference and generalizability. Simulation and physical robot experiments show the effectiveness of fine-grained 6-DoF grasping. These results show the promise of using the pre-trained language model, especially LLMs in embodied affordance reasoning and fine-grained grasping for a real robot.

\section{Acknowledgments}
\noindent This work acknowledges the support by the following programs: Postdoctoral Fellowship Program of CPSF (GZC20232292). National Natural Science Foundation of China (T2125009, 92048302). The funding of the "Pioneer" R\&D Program of Zhejiang (Grant No. 2023C03007).
Suzhou Key Laboratory of Artificial Intelligence and Social Governance Technologies (SZS2023007), Smart Social Governance Technology and Innovative Application Platform (YZCXPT2023101).


\bibliographystyle{ieeetr}



\end{document}